# A Surrogate-Assisted Variable Grouping Algorithm for General Large Scale Global Optimization Problems

An Chen, Zhigang Ren, Muyi Wang, Yongsheng Liang, Hanqing Liu, Wenhao Du

School of Automation Science and Engineering, Xi'an Jiaotong University, Xi'an, China

**Corresponding author name**: Zhigang Ren

**Affiliation**: School of Automation Science and Engineering, Xi'an Jiaotong University

**Permanent address**: No.28 Xianning West Road, Xi'an Shaanxi, 710049, P.R. China

**Email address**: renzg@xjtu.edu.cn

# A Surrogate-Assisted Variable Grouping Algorithm for General Large Scale Global Optimization Problems

An Chen, Zhigang Ren, Muyi Wang, Yongsheng Liang, Hanqing Liu, Wenhao Du

School of Automation Science and Engineering, Xi'an Jiaotong University, Xi'an, China

**Abstract:**

Problem decomposition plays a vital role when applying cooperative coevolution (CC) to large scale global optimization problems. However, most learning-based decomposition algorithms either only apply to additively separable problems or face the issue of false separability detections. Directing against these limitations, this study proposes a novel decomposition algorithm called surrogate-assisted variable grouping (SVG). SVG first designs a general-separability-oriented detection criterion according to whether the optimum of a variable changes with other variables. This criterion is consistent with the separability definition and thus endows SVG with broad applicability and high accuracy. To reduce the fitness evaluation requirement, SVG seeks the optimum of a variable with the help of a surrogate model rather than the original expensive high-dimensional model. Moreover, it converts the variable grouping process into a dynamic-binary-tree search one, which facilitates reutilizing historical separability detection information and thus reducing detection times. To evaluate the performance of SVG, a suite of benchmark functions with up to 2000 dimensions, including additively and non-additively separable ones, were designed. Experimental results on these functions indicate that, compared with six state-of-the-art decomposition algorithms, SVG possesses broader applicability and competitive efficiency. Furthermore, it can significantly enhance the optimization performance of CC.

## 1. Introduction

In the past decade, an increasing number of large scale global optimization (LSGO) problems have emerged from scientific research and engineering applications [13, 22, 23, 54]. Parameter training in deep neural networks [48], the design of potable water distribution networks [3], and petroleum reservoir management are a few typical examples [47]. Despite its prevalence, the LSGO problem is a hard nut to crack. For one thing, its mathematical model is difficult to build, making classic mathematical programming methods inapplicable. For another, evolutionary algorithms (EAs) suffer from the *curse of dimensionality* and lose their efficacy dramatically [2].

To relieve these difficulties, some researchers developed a *cooperative coevolution* (CC) architecture [5, 22, 28, 33, 37, 38, 50]. Taking the *divide-and-conquer* idea, CC first decomposes a complex LSGO problem into a set of simpler sub-problems and then cooperatively optimizes them using conventional EAs like differential evolution (DE) [5, 28, 50] and particle swarm optimization [1, 21]. As such, the problem decomposition plays a critical role in CC. It aims to group nonseparable variables into the same sub-problem and separable variables into different ones. A proper decomposition can decrease the solving difficulty of an LSGO problem without changing its optimum, but an improper one may lead CC to a Nash equilibrium rather than a real optimum [32]. Besides, it is expected that the decomposition process consumes as few fitness evaluations (FEs) as possible because even a single evaluation for an LSGO problem is generally very costly [39, 53].

Up to now, much research effort has been devoted to developing decomposition methods [22]. Nevertheless, some fundamental and pivotal challenges remain open. Early methods try to group variables statically [1, 33, 34] or randomly [21, 50, 52]. They perform well on fully separable problems but lose effectiveness on partially separable ones since they seldom consider the separability among variables [5, 28]. To overcome this drawback, learning-based methods perform decomposition by explicitly detecting variable separability. As a representative, differential grouping (DG) provides a simple separability detection criterion, i.e., checking whether the fitness variation caused by the perturbation on one variable is independent of other variables [28]. Since its inception, DG has attracted much research effort and a series of improved variants such as DG2 [31] and recursive DG (RDG) [41] were developed. Although these methods have achieved impressive decomposition performance, they only apply to additively separable problems. It is worth noting that some early learning-based methods such as variable interaction learning (VIL) [5] attempted to decompose general LSGO problems without exploiting characteristics of additive separability. VIL and its variants [9, 10, 43] group variables by detecting whether the monotonicity of the objective function w.r.t. a variable is affected by other ones and have the opportunity to identify non-additive separability. However, besides requiring a large number of FEs, these methods often encounter the issue of false separability detections [19].

To remedy the above shortcomings, this study proposes a surrogate-assisted variable grouping (SVG) algorithm. SVG designs *a general-separability-oriented detection criterion*. To determine the separability between a variable and other ones, this criterion checks whether the global optimum of this variable remains unchanged after perturbing the latter. It is consistent with the definition of general separability and thus applies well to generally separable problems. Moreover, this criterion only needs to locate the global optimum of a variable once. For a separable variable, the located optimum can be taken as its final optimum, which means that there is no need to tackle it again in the optimization process. Under the context of LSGO, the problem of locating the global optimum of a variable can be considered as a single-dimensional expensive optimization problem. SVG tries to solve it using the well-performed surrogate model technique [8, 14, 15, 39]. To be specific, it develops *a two-layer polynomial regression scheme*, which is able to obtain the global optimum of a variable with only a few FEs. As for the variable grouping process, SVG converts it into *a dynamic-binary-tree-based search one*. In this way, it systematically re-utilizes historical separability detection information and avoids lots of redundant detections. To verify the efficacy of SVG, we specially designed a more general suite of benchmark functions with up to 2000 dimensions and conducted comprehensive empirical studies on them. The results reveal the superiority of SVG over six state-of-the-art decomposition algorithms.

The main contributions of this study are as follows: 1) A new separability detection criterion is designed. It is consistent with the definition of general separability and possesses broader applicability than existing ones. 2) A surrogate-assisted scheme is introduced to seek the global optimum of a variable required by the new criterion. With the help of the surrogate technique, the global optimum of a variable can be located with a few FEs. 3) A dynamic-binary-tree-based variable grouping procedure is developed. It facilitates re-utilizing historical separability detection information and avoids lots of redundant detections. 4) A more general benchmark suite, including both additively and non-additively separable functions, is designed. The performance of SVG is comprehensively investigated on this benchmark suite. A thorough analysis of the empirical results is presented.

The remaining part of this paper proceeds as follows: Section 2 reviews the existing decomposition algorithms after briefly introducing the definition of separability and the framework of CC. Section 3 describes the proposed SVG algorithm in detail. Section 4 first introduces the designed benchmark suite and then reports experimental settings and results. Finally, Section 5 concludes this paper and discusses some future research directions.

## 2. Preliminaries

2.1 Separability and CC

**Definition 1.** A problem $f(x)$ is generally separable if it satisfies (assuming minimization) [45]:
$$\arg\min_{x} f(x) = (\arg\min_{x_1} f(\ldots, x_1, \ldots), \ldots, \arg\min_{x_k} f(\ldots, x_k, \ldots)), \quad (1)$$
where $x = (x_1, \ldots, x_n)$ is an $n$-dimensional decision vector and $x_1, \ldots, x_k$ are $k$ ($k = 2, \ldots, n$) disjoint subcomponents. We say the variables from the same subcomponent (or different subcomponents) to be non-separable (or separable) with each other. The variable separable with all the others is called a separable variable.

**Definition 2.** A problem $f(x)$ is additively separable if it has the following form [28]:
$$f(x) = \sum_{i=1}^{k} f_i(x_i), \quad (2)$$
where $f_i(\cdot)$ denotes the subfunction of the $i$-th nonseparable subcomponent.

It is obvious that additive separability is a special case of the general one. To further distinguish these two kinds of separability, let us compare the generally separable Ridge function $f(x_1, x_2) = \sqrt{x_1^2 + x_2^2}$ and the additively separable Sphere function $f(x_1, x_2) = x_1^2 + x_2^2$. For the latter, a perturbation on $x_1$ only causes the fitness landscape w.r.t. $x_2$ to translate. In other words, it does not affect the fitness variation caused by the change of $x_2$. By contrast, this property does not hold for the Ridge function whose separability only consists in that the global optimum of $x_2$ remains unchanged after perturbing $x_1$.

The difficulty of tackling an LSGO problem mainly lies in its large solution space that cannot be fully explored by traditional EAs. Two main categories of approaches have been developed to improve traditional EAs for LSGO, namely, CC methods and non-CC methods [11, 16, 18, 25, 26]. No-CC methods solve an LSGO problem as a whole and are generally equipped with some exploration-ability-enhanced operators [18, 25, 26] or some dimensionality reduction techniques [11, 16]. As a contrast, CC methods are developed by exploiting the separability of LSGO problems. **Algorithm 1** outlines the general framework of CC. It involves two main processes: decomposition and optimization. Note that since an explicit low-dimensional simulation model is unavailable for each sub-problem due to the black-box characteristic of the original LSGO problem, CC generally initializes a context vector (**cv**) with a complete solution to assist the evaluation of sub-solutions [1, 37] in line 2. Concretely, it inserts a sub-solution to be evaluated into the corresponding positions in **cv** and estimates its fitness by indirectly evaluating the modified **cv** with the original high-dimensional simulation model.

| **Algorithm 1**: Cooperative coevolution |
|---|
| 1. Perform decomposition: $x \to \{x_1, \ldots, x_k\}$; |
| 2. Initialize the context vector and the population for each sub-problem; |
| 3. **while** the termination criterion is not met |
| 4.      Determine the sub-problem $i$ to be optimized; |
| 5.      Optimize the $i$-th sub-problem for a certain number of generations; |
| 6.      Re-divide $x$ if necessary; |
| 7. **return** the best solution found so far. |

## 2.2 Decomposition methods

According to **Algorithm 1**, problem decomposition plays a fundamental role in CC. So far, lots of decomposition methods have been developed. The incipient decomposition methods are static ones. They directly divide an *n*-dimensional problem into *k* *m*-dimensional sub-problems ( $n = k \cdot m$ and $m << n$ ) and keep the grouping fixed during the following optimization process [1, 33, 34]. These methods are easy to implement but can only apply to fully separable problems due to the neglect of variable separability. Directing against the rigidness of static methods, random decomposition methods re-divide all decision variables in each evolution cycle [21, 50, 52]. The nonseparable variables thus have a chance of being assigned to the same sub-problem. Despite some improvement, random methods still perform poorly on the problems involving several groups of nonseparable variables [28].

Different from the above two types of decomposition methods, learning-based methods perform decomposition by explicitly detecting variable separability. Some studies attempted to learn variable separability by analyzing the solutions sampled in the optimization process. For instance, Ray and Yao [35] calculated the correlation coefficient between each pair of variables based on the current population individuals and assigned them to the same sub-problem if their correlation coefficient is larger than a predetermined threshold. Xu et al. [46] pointed out that this indicator cannot describe nonlinear separability and replaced it with mutual information. Omidvar et al. [30] detected variable separability according to the variation of each variable in two successive evolutionary cycles and thought that the variables with small changes are more likely to be nonseparable. By contrast, Yang et al. [49] tended to group the variables showing evolution consistency into the same sub-problem. In general, these decomposition methods can outperform static and random ones. However, as their separability detection criteria are set in a heuristic way, they can hardly achieve a proper decomposition.

To improve decomposition accuracy, some other studies designed separability detection criteria according to the characteristic of variable separability and judged whether the corresponding criteria hold by investigating the relationship among the fitness values of some purposefully sampled solutions. Inspired by the feature of additive separability, Omidvar et al. [28] developed the DG algorithm which takes the following detection criterion:

**Criterion 1**: For an *n*-dimensional function $f(\mathbf{x})$, its two variables $x_i$ and $x_j$ are additively separable if for $\forall \mathbf{cv} \in [\mathbf{lb}, \mathbf{ub}]^n$, $x_i^{'}, x_i^{"} \in [\mathbf{lb}(i), \mathbf{ub}(i)]$ and $x_j^{'}, x_j^{"} \in [\mathbf{lb}(j), \mathbf{ub}(j)]$ ( $x_i^{'} \neq x_i^{"}$ and $x_j^{'} \neq x_j^{"}$ ), the following condition holds:

$$\Delta(x_i^{'}, x_i^{"} | x_j^{'}) = \Delta(x_i^{'}, x_i^{"} | x_j^{"}),  \qquad (3)$$

where

$$\Delta(x_i^{'}, x_i^{"} | x_j^{'}) = f(\mathbf{cv} |\leftarrow x_i^{'}, x_j^{'}) - f(\mathbf{cv} |\leftarrow x_i^{"}, x_j^{'}) . \qquad (4)$$

Here, **lb** and **ub** represent the lower and upper bound of *x*, respectively, and $\mathbf{cv} |\leftarrow x_i^{'}, x_j^{'}$ denotes the complete solution generated by inserting $x_i^{'}$ and $x_j^{'}$ into the corresponding positions in **cv**.

Considering the computational roundoff error, DG introduces a predefined threshold $\varepsilon$ into Eq. (3) and converts it to $|\Delta(x_i^{'}, x_i^{"} | x_j^{'}) - \Delta(x_i^{'}, x_i^{"} | x_j^{"})| < \varepsilon$. DG achieves great success and is one of the most influential decomposition algorithms now. However, the original DG omits the indirect interdependency, i.e., the interdependency between two variables being directly separable but linked by other variables [42], and also faces the difficulty of setting a proper value for $\varepsilon$.

To identify indirect interdependency, some DG variants such as global DG (GDG) [24] and DG2 [31] detect the separability between each pair of variables and link each group of nonseparable variables together. Despite their effectiveness, these methods require $O(n^2)$ FEs for an *n*-dimensional problem. To reduce the FE consumption, some other DG variants detect separability from the perspective of variable subsets instead of variable individuals [4, 12, 17, 36, 41, 47]. The motivation behind this consists in that if two variable subsets $X_i$ and $X_j$ are separable, then each variable in $X_i$ is separable with the ones in $X_j$, thereby avoiding pairwise separability detection. The earliest variable-subset-based DG algorithm is fast interdependency identification (FII) [12]. It first excludes separable variables by investigating the separability between each variable and all the other ones and then further decomposes the remaining nonseparable variables. Different from FII, RDG [41] involves a recursive process, during which it equally divides the variable subset interacting with the current target variable and successively detects the separability between each resulting subset and the target variable. This process continues until all the variables interacting with the target variable are captured. By this means, RDG reduces the FE requirement to $O(n \log n)$. The efficient variable interdependency identification and decomposition method (EVIID) [17] further accelerates RDG by reutilizing some sampled solutions and pre-sorting variables according to the expected numbers of their expected nonseparable variables. Recently, the topology-based DG [47] algorithm integrates topological information into the decomposition process and remarkably improves decomposition efficiency.

As for the decomposition threshold $\varepsilon$, GDG [24] sets it to a value proportional to the minimum fitness of several randomly sampled solutions. Compared with the way to set $\varepsilon$ to a fixed value, this method provides certain adaptability and achieves higher decomposition accuracy. However, it still requires users to specify a proportion factor. To alleviate this issue, DG2 [31] first estimates the greatest lower bound and the least upper bound of the computational roundoff error and then configures $\varepsilon$ according to the estimation results. Instead of directly setting a value for $\varepsilon$, the study in [6] normalizes the indicator in DG and adaptively generates a threshold value for the indicator by analyzing its distribution.

Due to the additive feature of **Criterion 1**, DG and its variants only apply to additively separable problems. For generally separable ones, Chen et al. [5] proposed the well-known VIL algorithm. Its separability detection criterion can be detailed as follows:

**Criterion 2.** For an $n$-dimensional problem $f(\mathbf{x})$, $x_i$ and $x_j$ are separable if for $\forall \mathbf{cv} \in [\mathbf{lb}, \mathbf{ub}]^n$, $x_i', x_i'' \in [\mathbf{lb}(i), \mathbf{ub}(i)]$ and $x_j', x_j'' \in [\mathbf{lb}(j), \mathbf{ub}(j)]$ ($x_i' \neq x_i''$ and $x_j' \neq x_j''$), the following condition holds:

$$\Delta(x_i', x_i'' | x_j') \cdot \Delta(x_i', x_i'' | x_j'') > 0. \tag{5}$$

This criterion endows VIL with an ability to identify general separability among variables. However, for a pair of variables, it requires VIL to detect whether Eq. (5) holds many times, resulting in extremely high FE consumption. Even so, it was indicated that VIL may still omit some interdependencies [19]. Moreover, our study presented in Section 3.1 will reveal that VIL may also erroneously identify separable variables as nonseparable ones. To improve decomposition efficiency, some VIL variants extend **Criterion 2** by conducting separability detection from the perspective of variable subsets [9, 10]. For example, the fast variable interdependency searching (FVIS) [10] algorithm employs a recursive process like the one in RDG to perform decomposition and remarkably reduces the FE consumption. Nevertheless, it still cannot ensure a desirable decomposition accuracy.

Recently, Sun et al. [40] developed an another generally-separable-problem-oriented decomposition method called maximum entropic epistasis (MEE). It employs the maximal information coefficient to measure whether the partial derivative of a variable has a relationship with an another variable and identifies the separability between them accordingly. MEE adapts itself well to general separability but is conducted in a pairwise fashion and requires many evaluated samples to calculate the maximal information coefficient, which is unaffordable under the context of LSGO.

From previous studies, it can be known that a proper separability detection criterion is the basis to achieve accurate decomposition,

and reducing separability detection times and the number of FEs consumed in a single detection is the key to lower the FE requirement of a decomposition algorithm. Moreover, the study on decomposing generally separable LSGO problems is still unfolding. Given this situation, this study proposes the SVG algorithm. It can efficiently decompose generally separable LSGO problems not limited to additively separable ones.

## 3. Surrogate-Assisted Variable Grouping

### 3.1 A new general-separability-oriented detection criterion

To attain an accurate decomposition for a generally separable problem, a commensurate separability detection criterion is indispensable. Unfortunately, current predominant **Criterion 1** and **Criterion 2** show some apparent limitations, which can be intuitively illustrated by **Fig. 1**. **Figures 1(a)-(c)** present the fitness landscapes of $f(x_1,x_2)=x_1^2+x_2^2$, $f(x_1,x_2)=\sqrt{x_1^2+x_2^2}$, and $f(x_1,x_2)=x_1^2+x_2^2+x_1^2x_2^2e^{x_1x_2}$ w.r.t. $x_2$ before and after perturbing $x_1$. All these three functions are separable, where the first one is additively separable. However, **Criterion 1** only takes effect on **Fig. 1(a)** but fails to identify the separability shown by **Figs. 1(b)-(c)**. Take the case in **Fig. 1(b)** as an example. When $x_2$ is fixed to $x_2'$, the fitness variation caused by perturbing $x_1$ from 0 to 1 does not equal the corresponding variation when $x_2$ is fixed to $x_2''$, although the two fitness variations share the same sign. **Criterion 2** can successfully identify the separability shown by **Fig. 1(b)** but fails to tackle the case in **Fig. 1(c)**, where the two fitness variations have different signs. To this end, a more reliable separability detection criterion is required.

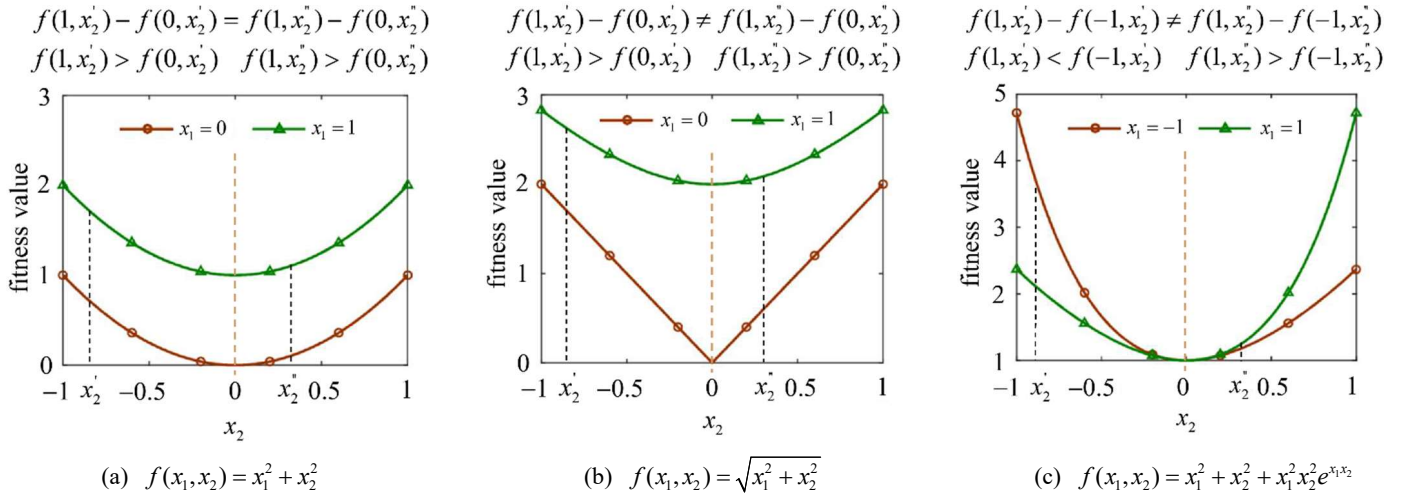

(a) $f(x_1,x_2)=x_1^2+x_2^2$     (b) $f(x_1,x_2)=\sqrt{x_1^2+x_2^2}$     (c) $f(x_1,x_2)=x_1^2+x_2^2+x_1^2x_2^2e^{x_1x_2}$

**Fig. 1.** Three examples to show the limitations of **Criterion 1** and **Criterion 2**.

According to **Definition 1**, a target variable $x_t$ is generally separable with an undetected one $x_u$ if its global optimum is independent of the latter. A direct approach to detecting this hallmark is to check whether the two global optima of $x_t$ obtained before and after perturbing $x_u$ equal each other. However, it needs to locate the global optimum of $x_t$ twice in a single detection and thus consumes excessive FEs. To avoid this situation, a simple way is to check whether the previous optimum of $x_t$ still holds after perturbing $x_u$. This basic idea can be illustrated by **Fig. 2**, where each curve represents the fitness landscape w.r.t. $x_t$ around its global optimum. Let $x_t^*$ denote the previous global optimum of $x_t$. Then as shown in **Fig. 2(a)**, if $x_t$ and $x_u$ are generally separable, $x_t^*$ keeps its optimality after perturbing $x_u$, which means that the fitness value of $x_t^*$ is still smaller than those of $x_t^*-\delta$ and $x_t^*+\delta$ with $\delta$ being a small positive number. Otherwise, the global optimum of $x_t$ will move, which means that the fitness value of $x_t^*$ must be larger than that of $x_t^* - \delta$ (as shown in **Fig. 2(b)**) or $x_t^* + \delta$ (as shown in **Fig. 2(c)**). By summarizing the cases shown in **Figs. 2(a)-(c)**, we can get a simple yet efficient separability detection criterion as follows:

**Criterion 3**: For an *n*-dimensional function $f(x)$ and $\forall \mathbf{cv} \in [\mathbf{lb}, \mathbf{ub}]^n$, suppose the current global optimum of a variable $x_t$ is $x_t^*$. Then $x_t$ is generally separable with an another variable $x_u$ if the following condition holds:

$$f(\mathbf{cv} \,|\!\leftarrow x_t^*, x_u^{'}) < \min\{f(\mathbf{cv} \,|\!\leftarrow x_t^* - \delta, x_u^{'}), f(\mathbf{cv} \,|\!\leftarrow x_t^* + \delta, x_u^{'})\}, \tag{6}$$

where $x_u^{'}$ is a value of $x_u$ satisfying $x_u^{'} \in [\mathbf{lb}(u), \mathbf{ub}(u)]$ and $x_u^{'} \neq \mathbf{cv}(u)$. Otherwise, the two variables are generally nonseparable.

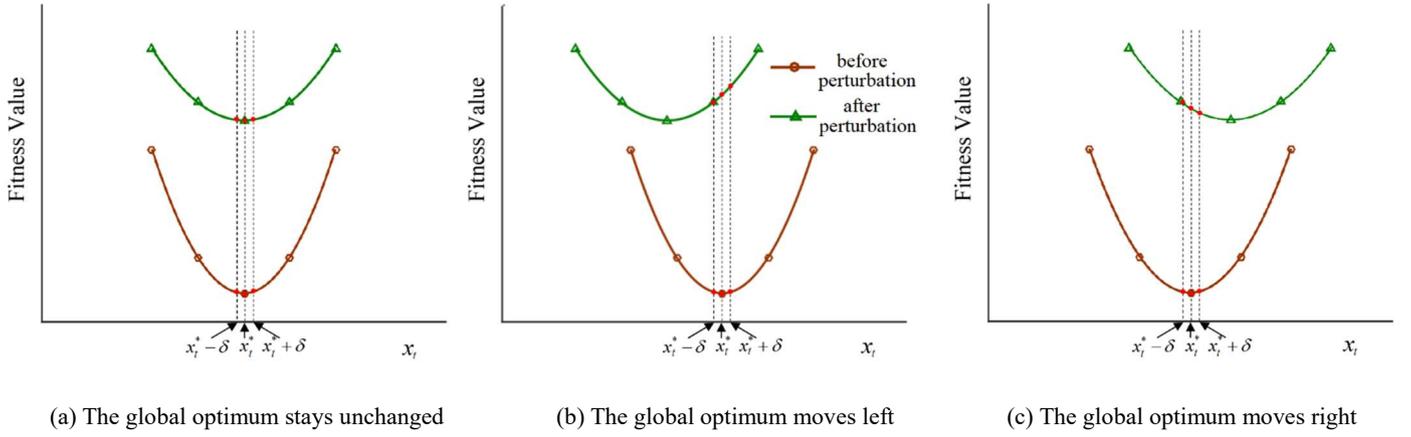

(a) The global optimum stays unchanged  (b) The global optimum moves left  (c) The global optimum moves right

**Fig. 2.** The basic idea of **Criterion 3**.

This criterion possesses some superiorities over **Criterion 1** and **Criterion 2**. It extends **Criterion 1** by applying to non-additive separability. It can also avoid the false detection encountered by **Criterion 2**. Besides, some further remarks are deserved to make on **Criterion 3**. Firstly, like **Criterion 1** and **Criterion 2**, the undetected variable $x_u$ can be replaced with a variable subset $X_u$. If the optimum of $x_t$ is independent with $X_u$, then $x_t$ can be judged separable with each variable in $X_u$. Otherwise, there must be one or more variables in $X_u$ interacting with $x_t$. Secondly, this criterion requires to locate $x_t^*$ in advance. This is a seemingly challenging problem, considering the FEs required by an optimization process. Fortunately, despite its expensive characteristic, this optimization problem merely involves a single variable and can be efficiently solved by the surrogate model technique [8, 14, 15, 39]. Moreover, if $x_t$ is a separable variable, the located $x_t^*$ can be directly taken as its final optimum. Therefore, we need not tackle it again in the optimization process of CC. Finally, $\delta$ specifies a small neighborhood of $x_t^*$. We may set a value for it according to the step size when searching for $x_t^*$. The details will be described in the next subsection.

**Algorithm 2**: $(isSep, FEsUsed) \leftarrow \text{DetectSep}(x_t, X_u, x_t^*, \delta, \mathbf{cv}, \mathbf{cv}^{'})$

1. Perform initializations: $FEsUsed \leftarrow 0$, $isSep \leftarrow flase$, and $\mathbf{s} \leftarrow \mathbf{cv}$;
2. **For** each $x_u$ in $X_u$
3. $\quad$ Set $\mathbf{s}(u)$ to $\mathbf{cv}^{'}(u)$;
4. Set $\mathbf{s}_1$, $\mathbf{s}_2$, and $\mathbf{s}_3$: $\mathbf{s}_1, \mathbf{s}_2, \mathbf{s}_3 \leftarrow \mathbf{s}$ and $\mathbf{s}_1(t) \leftarrow x_t^* - \delta$, $\mathbf{s}_2(t) \leftarrow x_t^*$, $\mathbf{s}_3(t) \leftarrow x_t^* + \delta$;
5. **if** $f(\mathbf{s}_2) < \min\{f(\mathbf{s}_1), f(\mathbf{s}_3)\}$ **then**
6. $\quad$ $isSep \leftarrow true$;
7. Update $FEsUsed$: $FEsUsed \leftarrow FEsUsed + 3$;
8. **return** $isSep$ and $FEsUsed$.

**Algorithm 2** presents the pseudocode of applying **Criterion 3** to detecting the separability between $x_t$ and $X_u$, where **cv**' is employed to specify the value of $X_u$ after perturbation. For simplicity, we set **cv** and **cv**' to **lb** and $(\mathbf{lb}+\mathbf{ub})/2$, respectively.

3.2 A two-layer polynomial regression scheme

As described above, we need to seek the global optimum of a variable before applying **Criterion 3**. An EA may effortlessly accomplish this task but generally requires many FEs. To reduce the FE requirement, this study employs the well-performed surrogate model technique to tackle this issue [8, 14, 15, 39]. Its main idea is to first construct a surrogate model for the original problem using some real-evaluated solutions, and then to efficiently and approximatively evaluate candidate solutions with the constructed model. As a consequence, many real FEs can be avoided. Up to now, several types of classic surrogate models [14, 15], including polynomial regression (PR), Gaussian process, and radial basis function, have been developed. Considering the one-dimension and possible multi-model features of the problem to be optimized, this study designs a *two-layer polynomial regression* (TLPR) *scheme*.

As its name implies, TLPR involves a bilayer surrogate structure. The first layer is about a global PR model. It aims to approximate the global profile of the fitness landscape by smoothing out the local optima. With its help, a promising region covering the global optimum of the problem is expected to be obtained. Then in the second layer, TLPR first divides the obtained trust region into several segments and then constructs a local PR model for each one. As each local model is just responsible for a small segment, it can achieve very high accuracy. By comparing the optima provided by all the local PR models, an expectant global optimum can be obtained.

The basic PR model for a one-dimensional problem can be formulated as follows:

$$PR(x) = p_r x^r + p_{r-1} x^{r-1} + \cdots + p_0 x^0, \tag{7}$$

where $p_r, p_{r-1}, \ldots, p_0$ are the coefficients and can be approximated by applying the least-squares training method based on at least $r+1$ samples. The degree $r$ determines the expression preference of the built model. The model of a low degree (e.g. two) can smooth out the local optima of the problem and thus is good at capturing its global profile. By contrast, the model of a high degree (e.g. five) can describe fitness features in detail but may over fit. Motivated by these properties, TLPR employs the two-degree and five-degree PR models in the first and second layers, respectively. Note that the global optima of these two kinds of PR models can be directly deduced according to their analytical expressions.

**Algorithm 3** describes the pseudocode of TLPR. Lines 1-4 are about the first layer, where a two-degree PR model is constructed based on 100 uniformly sampled solutions. As our preliminary study suggested, such a configuration can provide enough solutions around the global optimum and thus approximate the global profile of the fitness landscape well. After locating the optimum of the two-degree PR model in line 3, **Algorithm 3** takes it as the center of the trust region and specifies the region as 10% of the feasible one in line 4. Lines 5-9 are about the second layer. To divide the trust region as finely as possible, **Algorithm 3** takes each six sample solutions, the minimum number required for training a five-degree PR model, as a unit to construct the model in lines 7-8. For the preliminary global optimum $x_t^*$ chosen in line 9, line 10 further improves it with a local search procedure based on the original simulation model. The classic BFGS method is employed here for its impressive performance [27]. As $x_t^*$ is near the real global optimum, BFGS generally gets converged within several iterations. Its final step size is also output and can be naturally employed to specify the small neighborhood $\delta$ for **Criterion 3**.

**Algorithm 3**: $(x_t^*, \delta, FEsUsed) \leftarrow \text{TLPR}(x_t, \mathbf{lb}, \mathbf{ub}, \mathbf{cv})$

1. Uniformly sample 100 solutions in $[\mathbf{lb}(t), \mathbf{ub}(t)]$;
2. Evaluate the sampled solutions based on **cv** and set *FEsUsed* to 100;
3. Construct a two-degree PR model and locate its optimum $x_t^{*'}$;
4. Define the trust region as 10% of $[\mathbf{lb}(t), \mathbf{ub}(t)]$ centered on $x_t^{*'}$;
5. Uniformly re-sample another 100 solutions in the trust region;
6. Evaluate the sampled solutions and update *FEsUsed*: $FEsUsed \leftarrow FEsUsed + 100$;
7. **For** each six sampled solutions
8.     Construct a five-degree PR model and locate the candidate optimum;
9. Pick out the best optimum $x_t^*$ among all the candidates;
10. Conduct BFGS to further improve $x_t^*$;
11. Set $\delta$ to the last step size of BFGS and update FEs;
12. **return** $x_t^*$, $\delta$, and *FEsUsed*.

3.3 A dynamic-binary-tree-based variable grouping procedure

The key of problem decomposition lies in that, for the current target variable $x_t$, how to efficiently extract concrete variables interacting with it from the undetected variable set $X_u$. A direct way is to check the separability between $x_t$ and each variable individual in $X_u$ one by one. However, it requires a considerable number of FEs. Directing against the inefficiency, we develop a dynamic-binary-tree-based variable grouping (DBTG) procedure.

DBTG works as follows: it first sets $X_u$ to be the root node and detects its separability with $x_t$. If they interact with each other, DBTG equally divides $X_u$ into two child nodes and successively detects their separability with $x_t$. A node will not be further divided if it only involves a single variable or is separable with $x_t$. DBTG shares some similarities with the recursive process in RDG [41], which can be also described by a binary tree. Besides taking a different separability detection criterion, another improvement of DBTG lies in that it reduces separability detection times by reutilizing some historical detection information. To get a better understanding of this mechanism, let us consider the following example:

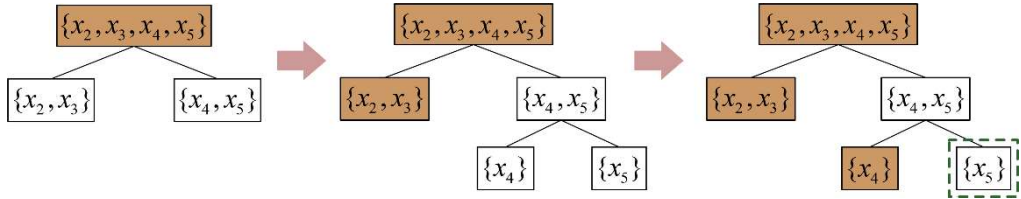

**Fig. 3.** An example for illustrating the work process of DBTG.

*Example*: For the problem $f(\mathbf{x}) = (x_1 - x_5)^2 + (x_1 - 1)^2 + x_2^2 + x_3^2 + x_4^2$, **Fig. 3** presents the process of DBTG when capturing the variables interacting with $x_1$. After identifying the nonseparability between $x_1$ and $\{x_2, x_3, x_4, x_5\}$, DBTG equally divides the latter into two child nodes and then detects the separability between $\{x_2, x_3\}$ and $x_1$. Since they are separable, it can be deduced that $\{x_4, x_5\}$ is nonseparable with $x_1$. The reason behind this deduction is that a variable subset nonseparable with $x_1$ must cover at least one variable interacting with it. Thus DBTG directly divides $\{x_4, x_5\}$ into $\{x_4\}$ and $\{x_5\}$. Similarly, after identifying the separability between $x_1$ and $\{x_4\}$, DBTG judges $x_5$ to be nonseparable with $x_1$ without further detection. This means that compared with the existing recursive decomposition process, DBTG avoids two unnecessary detections.

**Algorithm 4** presents the pseudocode of DBTG. A queue *treeNodes* is set to dynamically store the nodes interacting with $x_t$. For each node popped out from *treeNodes*, if it involves a single variable, **Algorithm 4** directly adds it to $X_t^{'}$ in lines 8-9 which stores all the variables interacting with $x_t$. Otherwise, it halves this node into two child ones in line 11. After detecting the separability between the first child node and $x_t$ in lines 12-13, DBTG further detects or deduces the separability between the other one and $x_t$ in lines 16-19, and push the two nodes into *treeNodes* if they are nonseparable with $x_t$. The process will not terminate until *treeNodes* becomes empty.

---

**Algorithm 4**: $(X_t^{'}, FEsUsed) \leftarrow \text{DBTG}(x_t, X_u, x_t^*, \delta, \mathbf{cv}, \mathbf{cv}^{'})$

1. Perform initializations: $X_t^{'} \leftarrow \varnothing$, $FEsUsed \leftarrow 0$, and $treeNodes \leftarrow \varnothing$;
2. Detect the separability between $x_t$ and $X_u$:
   $(isSep, FEsUsed^{'}) \leftarrow \text{DetectSep}(x_t, X_u, x_t^*, \delta, \mathbf{cv}, \mathbf{cv}^{'})$;
3. Update $FEsUsed$: $FEsUsed \leftarrow FEsUsed + FEsUsed^{'}$;
4. **if** $isSep == false$ **then**
5.  Push $X_u$ into the rear of *treeNodes*;
6.  **while** $|treeNodes| > 0$
7.   Pop out the first node of *treeNodes* and denote it as $X_c$;
8.   **if** $|X_c| == 1$ **then**
9.    Update $X_t^{'}$: $X_t^{'} \leftarrow X_t^{'} \cup X_c$;
10.   **else**
11.    Equally divide $X_c$: $X_c \rightarrow X_{c1} \cup X_{c2}$ and set $isSep2$ to *false*;
12.    Detect the separability between $x_t$ and $X_{c1}$:
    $(isSep1, FEsUsed^{'}) \leftarrow \text{DetectSep}(x_t, X_{c1}, x_t^*, \delta, \mathbf{cv}, \mathbf{cv}^{'})$;
13.    Update $FEsUsed$: $FEsUsed \leftarrow FEsUsed + FEsUsed^{'}$;
14.    **if** $isSep1 == false$ **then**
15.     Push $X_{c1}$ into the rear of *treeNodes*;
16.    Detect the separability between $x_t$ and $X_{c2}$:
    $(isSep2, FEsUsed^{'}) \leftarrow \text{DetectSep}(x_t, X_{c2}, x_t^*, \delta, \mathbf{cv}, \mathbf{cv}^{'})$;
17.    Update $FEsUsed$: $FEsUsed \leftarrow FEsUsed + FEsUsed^{'}$;
18.    **if** $isSep2 == false$ **then**
19.     Push $X_{c2}$ into the rear of *treeNodes*;
20. **return** $X_t^{'}$ and $FEsUsed$.

---

Taking DBTG as a key component, **Algorithm 5** presents the pseudocode of the whole SVG algorithm. After performing initializations, it starts an iteration by randomly selecting a variable from the undetected variable subset $X_u$ as the target variable $x_t$ in line 4. Then it successively locates the global optimum of $x_t$ using the proposed TLPR scheme in line 5 and extracts the concrete variables interacting with $x_t$ by performing DBTG in line 9. If $x_t$ is separable, it is stored into *seps* in lines 8 and 12; Otherwise, $x_t$ together with its partners is stored into *nonseps* in line 14. SVG iteratively conducts the above process until all the variables are tackled. Note that once the global optimum of a variable is obtained, it is employed to update the corresponding component in **cv**, which will be further transferred to the optimization process of CC and thus provides a promising initial solution. Since the global optima of separable variables have been obtained, we only need to focus on nonseparable variables in the optimization process.

**Algorithm 5:** $(seps, nonseps, \mathbf{cv}, FEsUsed) \leftarrow \text{SVG}(\mathbf{lb}, \mathbf{ub})$

1. Perform initializations: $seps, nonseps \leftarrow \varnothing$, $\mathbf{cv} \leftarrow \mathbf{lb}$, $\mathbf{cv}' \leftarrow (\mathbf{lb}+\mathbf{ub})/2$ and $FEsUsed \leftarrow 0$;
2. Assign all decision variables to $X_u$;
3. **while** $|X_u| > 1$
4.     Initialize $x_t$ with a random variable in $X_u$ and delete it from $X_u$;
5.     Locate the global optimum of $x_t$: $(x_t^*, \delta, FEsUsed') = \text{TLPR}(x_t, \mathbf{lb}, \mathbf{ub}, \mathbf{cv})$;
6.     Update $\mathbf{cv}$ and $FEsUsed$: $\mathbf{cv}(t) \leftarrow x_t^*$ and $FEsUsed \leftarrow FEsUsed + FEsUsed'$;
7.     **if** $|X_u| == 0$ **then**
8.        Update $seps$: $seps \leftarrow seps \cup \{x_t\}$; **break**;
9.     Extract the variables interacting with $x_t$: $(X_t', FEsUsed') \leftarrow \text{DBTG}(x_t, X_u, x_t^*, \delta, \mathbf{cv}, \mathbf{cv}')$;
10.    Update $FEsUsed$: $FEsUsed \leftarrow FEsUsed + FEsUsed'$;
11.    **if** $|X_t'| == 0$ **then**
12.       Update $seps$: $seps \leftarrow seps \cup \{x_t\}$;
13.    **else**
14.       Update $nonseps$ and $X_u$: $nonseps \leftarrow nonseps \cup \{x_t \cup X_t'\}$ and $X_u \leftarrow X_u \setminus X_t'$;
15. **return** $seps$, $nonseps$, $\mathbf{cv}$ and $FEsUsed$.

## 3. Experimental studies

In this section, we first introduce the new benchmark suite designed for generally separable LSGO problems, and then sequentially evaluate the performance of SVG in terms of decomposition accuracy and efficiency, capability in enhancing the optimization performance of CC, and scalability by comparing it with six state-of-the-art decomposition methods on the new benchmark suite.

### 4.1 The new benchmark suite

To date, several LSGO benchmark suites have been developed and played important roles in evaluating the performance of LSGO algorithms [20, 29, 44]. Nevertheless, the separable functions in these suites are generally limited to additively separable ones, which cannot completely reflect the features of separable problems in practice and may mislead the development of decomposition algorithms. For example, the widely-used CEC'2010 benchmark suite involves 18 fully or partially separable functions, 15 out of which are additively separable ones [44]. To this end, we design a more general benchmark suite in this study.

The new benchmark suite takes a similar function generation method as the CEC'2010 suite [44], but introduces two more generally separable functions, i.e., the Exponential function and the Ridge function, as basic ones to generate new benchmark functions. All the involved basic functions are listed as follows:

1) The *n*-dimensional additively separable Elliptic Function:

$$f_{elli}(\mathbf{x}) = \sum_{i=1}^{n}(10^6)^{\frac{i-1}{n-1}} x_i^2, \quad \mathbf{x} \in [-100,100]^n;$$

2) The *n*-dimensional additively separable Rastrigin's Function:

$$f_{rast}(\mathbf{x}) = \sum_{i=1}^{n}[x_i^2 - 10\cos(2\pi x_i) + 10], \quad \mathbf{x} \in [-5,5]^n;$$

3) The *n*-dimensional generally separable Exponential Function:

$$f_{expo}(\mathbf{x}) = 200 - 200 \cdot exp(-\frac{1}{n} \cdot \sqrt{\sum_{i=1}^{n} \frac{i}{n} x_i^2}), \quad \mathbf{x} \in [-32,32]^n;$$

4) The *n*-dimensional generally separable Ackley Function:

$$f_{ackl}(x) = -20 \cdot \exp\left(\sqrt{\frac{1}{n}\sum_{i=1}^{n} x_i^2}\right) - \exp\left(\frac{1}{n}\sum_{i=1}^{n} \cos(2\pi x_i)\right) + 20 + e, \quad x \in [-32, 32]^n;$$

5) The *n*-dimensional generally separable Ridge Function:

$$f_{ridg}(x) = n \cdot \sqrt{\sum_{i=1}^{n} x_i^2}, \quad x \in [-100, 100]^n;$$

6) The *n*-dimensional nonseparable Schwefel Function:

$$f_{schw}(x) = \sum_{i=1}^{n} \left(\sum_{j=1}^{i} x_i\right)^2, \quad x \in [-100, 100]^n.$$

**Table 1** presents the details of the generated 21 benchmark functions, which can be divided into five categories, including fully separable functions, partially separable ones, and nonseparable ones. The symbol **o** denotes the global optimum satisfying $f(\mathbf{o}) = 0$, $z$ represents the translation of a candidate solution $x$ from **o**, the subscript *Rot* denotes the coordinate rotation operator making the corresponding variables nonseparable, and *m* specifies the size of a nonseparable variable subcomponent. The problem dimension $n$ and $m$ are set to 1000 and 50, respectively. Overall, nine functions, including $f_3$-$f_5$, $f_8$-$f_{10}$, and $f_{13}$-$f_{15}$, are generally separable but are not additively separable.

**Table 1.** Details of the 21 new benchmark functions.

| Func. | Equation | Categories |
|---|---|---|
| $f_1$ | $f(x) = f_{elli}(z), \; z = x - \mathbf{o}$ | |
| $f_2$ | $f(x) = f_{rast}(z)$ | |
| $f_3$ | $f(x) = f_{expo}(z)$ | Fully separable function |
| $f_4$ | $f(x) = f_{ackl}(z)$ | |
| $f_5$ | $f(x) = f_{ridg}(z)$ | |
| $f_6$ | $f(x) = f_{elli}(z_{Rot}(1:m)) + f_{elli}(z(m+1:n))$ | |
| $f_7$ | $f(x) = f_{rast}(z_{Rot}(1:m)) + f_{rast}(z(m+1:n))$ | Partially separable function with a single $m$-dimensional subcomponent |
| $f_8$ | $f(x) = f_{expo}(z_{Rot}(1:m)) + f_{expo}(z(m+1:n))$ | |
| $f_9$ | $f(x) = f_{ackl}(z_{Rot}(1:m)) + f_{ackl}(z(m+1:n))$ | |
| $f_{10}$ | $f(x) = f_{schw}(z(1:m)) + f_{ridg}(z(m+1:n))$ | |
| $f_{11}$ | $f(x) = \sum_{i=1}^{n/(2m)} f_{elli}(z_{Rot}((i-1)*m+1:i*m)) + f_{elli}(z(n/2+1:n))$ | |
| $f_{12}$ | $f(x) = \sum_{i=1}^{n/(2m)} f_{rast}(z_{Rot}((i-1)*m+1:i*m)) + f_{rast}(z(n/2+1:n))$ | Partially separable function with $n/(2m)$ $m$-dimensional subcomponents |
| $f_{13}$ | $f(x) = \sum_{i=1}^{n/(2m)} f_{expo}(z_{Rot}((i-1)*m+1:i*m)) + f_{expo}(z(n/2+1:n))$ | |
| $f_{14}$ | $f(x) = \sum_{i=1}^{n/(2m)} f_{ackl}(z_{Rot}((i-1)*m+1:i*m)) + f_{ackl}(z(n/2+1:n))$ | |
| $f_{15}$ | $f(x) = \sum_{i=1}^{n/(2m)} f_{schw}(z((i-1)*m+1:i*m)) + f_{ridg}(z(n/2+1:n))$ | |
| $f_{16}$ | $f(x) = \sum_{i=1}^{n/m} f_{elli}(z_{Rot}((i-1)*m+1:i*m))$ | |
| $f_{17}$ | $f(x) = \sum_{i=1}^{n/m} f_{rast}(z_{Rot}((i-1)*m+1:i*m))$ | Partially separable function with $n/m$ $m$-dimensional subcomponents |
| $f_{18}$ | $f(x) = \sum_{i=1}^{n/m} f_{expo}(z_{Rot}((i-1)*m+1:i*m))$ | |
| $f_{19}$ | $f(x) = \sum_{i=1}^{n/m} f_{ackl}(z_{Rot}((i-1)*m+1:i*m))$ | |
| $f_{20}$ | $f(x) = \sum_{i=1}^{n/m} f_{schw}(z((i-1)*m+1:i*m))$ | |
| $f_{21}$ | $f(x) = f_{schw}(z)$ | Nonseparable function |

## 4.2 Investigation of decomposition performance

To reveal the decomposition performance of SVG, we experimentally compared it with six state-of-the-art decomposition algorithms, including DG, DG2, RDG, EVIID, VIL, and FVIS, whose main ideas have been briefly introduced in Section 2.2. For a fair comparison, the parameters involved in each competitor were set according to the suggestions in the corresponding original paper. The decomposition efficiency and accuracy of each algorithm were quantified by the number of consumed FEs and an indicator named normalized mutual information (NMI), respectively. The latter is defined as follows [7]:

**Definition 3.** For an *n*-dimensional problem $f(x)$, suppose $D = \{X_1, \ldots, X_{k_1}\}$ and $D' = \{X'_1, \ldots, X'_{k_2}\}$ are the ideal decomposition and the one generated by a decomposition algorithm, respectively. The NMI indicator between $D$ and $D'$ can be formulated as

$$\text{NMI}(D, D') = \frac{-2 \times \sum_{i=1}^{k_1} \sum_{j=1}^{k_2} M_{ij} \log_2 \left( nM_{ij} / |X_i| \cdot |X'_j| \right)}{\sum_{i=1}^{k_1} |X_i| \log_2 \left( |X_i|/n \right) + \sum_{j=1}^{k_2} |X'_j| \log_2 \left( |X'_j|/n \right)} \times 100\%, \quad (8)$$

where $M$ denotes the confusion matrix with each element $M_{ij}$ representing the number of common variables in $X_i$ and $X'_j$. This metric is derived from mutual information with its value varying in the range $[0,1]$. NMI can precisely quantify the consistency between $D$ and $D'$. The more consistent they are, the larger $\text{NMI}(D, D')$ is. Specially, if $D'$ is the same as $D$, then $\text{NMI}(D, D') = 1$.

**Table 2** lists the decomposition results of SVG and its six competitors on the developed benchmark suite. For a more in-depth analysis of the decomposition accuracy of each algorithm, we separately calculated its NMI values on the separable and nonseparable variables, which are denoted by $\rho_1$ and $\rho_2$ in **Table 2**, respectively. For SVG, the entry *dis* shows the Euclidean distance between the subcomponent of the final **cv** w.r.t. the identified separable variables and the corresponding real optima, and thus can quantify the performance of the developed TLPR scheme. The bottom row of **Table 2** averages the results of each algorithm on all the functions. According to **Table 2**, the following comparative analyses can be made:

1) *Decomposition accuracy*: For separable variables, SVG achieves 100% decomposition accuracy ($\rho_1$) on 12 out of all the 15 functions containing separable variables and nearly 100% accuracy on the other three functions, i.e., $f_4$, $f_9$, and $f_{14}$. These three functions are all the variants of the Ackley function, where each decision variable involves many local optima. The imperfect of SVG on them comes from that the TLPR scheme occasionally mistakes the local optima of some target variables as the global ones, leading to improper decompositions. As discussed above, DG and its variants achieve satisfying accuracy on additively separable functions but lose their competitiveness on most non-additively separable ones. For example, DG2, RDG, and EVIID all perform poorly on $f_3$-$f_5$ and $f_{13}$-$f_{15}$. However, it is somewhat counterintuitive that DG achieves 100% $\rho_1$ on $f_3$ and $f_4$, RDG achieves 100% $\rho_1$ on $f_9$, and EVIID achieves 100% $\rho_1$ on $f_8$-$f_{10}$. A closer inspection reveals that their success on these non-additively separable functions can be attributed to large threshold values, which unintentionally make **Criterion 1** correctly identify general separability. As for VIL and FVIS, they correctly identify most separable variables, but also make some mistakes on the variants of the Ackley function.

**Table 2.** Decomposition results of the seven decomposition algorithms on 1000-dimensional benchmark functions.

| Func. | DG $\rho_1$ | DG $\rho_2$ | DG FEsUsed | DG2 $\rho_1$ | DG2 $\rho_2$ | DG2 FEsUsed | RDG $\rho_1$ | RDG $\rho_2$ | RDG FEsUsed | EVIID $\rho_1$ | EVIID $\rho_2$ | EVIID FEsUsed | VIL $\rho_1$ | VIL $\rho_2$ | VIL FEsUsed | FVIS $\rho_1$ | FVIS $\rho_2$ | FVIS FEsUsed | SVG $\rho_1$ | SVG $\rho_2$ | SVG FEsUsed | dis |
|---|---|---|---|---|---|---|---|---|---|---|---|---|---|---|---|---|---|---|---|---|---|---|
| $f_1$ | 100 | — | 1.00e+06 | 100 | — | 5.01e+05 | 100 | — | 3.01e+03 | 100 | — | 3.01e+03 | 100 | — | 6.60e+04 | 100 | — | 4.00e+04 | 100 | — | 2.12e+05 | 0.00e+00 |
| $f_2$ | 100 | — | 1.00e+06 | 100 | — | 5.01e+05 | 100 | — | 3.01e+03 | 100 | — | 3.01e+03 | 100 | — | 6.60e+04 | 100 | — | 4.00e+04 | 100 | — | 2.10e+05 | 1.63e–14 |
| $f_3$ | 100 | — | 1.00e+06 | 0.00 | — | 5.01e+05 | 0.00 | — | 6.01e+03 | 0.00 | — | 4.01e+03 | 100 | — | 6.60e+04 | 100 | — | 4.00e+04 | 100 | — | 2.10e+05 | 3.32e–06 |
| $f_4$ | 100 | — | 1.00e+06 | 0.00 | — | 5.01e+05 | 0.00 | — | 6.00e+03 | 0.00 | — | 4.09e+03 | 99.42 | — | 1.80e+06 | 100 | — | 4.44e+04 | 99.65 | — | 1.99e+05 | 1.30e–10 |
| $f_5$ | 0.00 | — | 2.00e+03 | 0.00 | — | 5.01e+05 | 0.00 | — | 6.00e+03 | 0.00 | — | 4.01e+03 | 100 | — | 6.60e+04 | 100 | — | 4.00e+04 | 100 | — | 2.02e+05 | 8.49e–13 |
| $f_6$ | 41.92 | 100 | 1.33e+04 | 100 | 100 | 5.01e+05 | 100 | 100 | 4.20e+03 | 100 | 100 | 3.08e+03 | 99.89 | 0.00 | 1.80e+06 | 100 | 0.00 | 3.02e+05 | 100 | 100 | 2.03e+05 | 0.00e+00 |
| $f_7$ | 100 | 100 | 9.07e+05 | 100 | 100 | 5.01e+05 | 100 | 100 | 4.23e+03 | 100 | 100 | 3.08e+03 | 99.62 | 0.00 | 1.80e+06 | 100 | 0.00 | 5.39e+04 | 100 | 100 | 2.02e+05 | 1.62e–14 |
| $f_8$ | 100 | 100 | 9.06e+05 | 2.85 | 100 | 5.01e+05 | 8.08 | 100 | 9.12e+03 | 100 | 100 | 3.77e+04 | 100 | 0.00 | 6.60e+04 | 100 | 0.00 | 1.43e+05 | 100 | 100 | 1.93e+05 | 1.55e–07 |
| $f_9$ | 100 | 100 | 9.08e+05 | 15.30 | 100 | 5.01e+05 | 100 | 100 | 5.08e+04 | 100 | 100 | 5.97e+03 | 97.57 | 0.00 | 1.80e+06 | 100 | 100 | 5.25e+04 | 99.69 | 100 | 1.91e+05 | 1.26e–10 |
| $f_{10}$ | 1.70 | 100 | 2.15e+03 | 30.87 | 100 | 5.01e+05 | 0.72 | 100 | 8.43e+03 | 100 | 100 | 7.37e+04 | 100 | 100 | 1.80e+06 | 100 | 0.00 | 1.95e+05 | 100 | 100 | 1.94e+05 | 3.25e–12 |
| $f_{11}$ | 100 | 100 | 2.70e+05 | 100 | 100 | 5.01e+05 | 100 | 100 | 1.39e+04 | 100 | 100 | 7.66e+03 | 98.17 | 38.62 | 1.80e+06 | 100 | 60.35 | 1.40e+06 | 100 | 100 | 1.23e+05 | 0.00e+00 |
| $f_{12}$ | 100 | 100 | 2.71e+05 | 100 | 100 | 5.01e+05 | 100 | 100 | 1.43e+04 | 100 | 100 | 8.74e+03 | 99.18 | 54.40 | 1.80e+06 | 100 | 97.70 | 1.25e+05 | 100 | 100 | 1.21e+05 | 1.14e–14 |
| $f_{13}$ | 91.83 | 97.97 | 1.75e+05 | 0.00 | 100 | 5.01e+05 | 0.00 | 100 | 1.35e+04 | 0.00 | 100 | 8.90e+03 | 99.66 | 59.32 | 1.81e+06 | 100 | 55.38 | 9.29e+05 | 100 | 100 | 1.16e+05 | 1.79e–06 |
| $f_{14}$ | 100 | 99.79 | 2.85e+05 | 0.00 | 100 | 5.01e+05 | 0.00 | 100 | 1.35e+04 | 0.00 | 100 | 6.97e+03 | 98.05 | 51.15 | 1.81e+06 | 100 | 100 | 1.09e+05 | 98.48 | 100 | 1.10e+05 | 5.43e–11 |
| $f_{15}$ | 0.00 | 100 | 9.30e+03 | 0.00 | 100 | 5.01e+05 | 0.00 | 100 | 1.33e+04 | 0.00 | 100 | 7.50e+03 | 100 | 89.79 | 1.81e+06 | 100 | 64.75 | 1.67e+06 | 100 | 100 | 1.16e+05 | 4.83e–13 |
| $f_{16}$ | — | 100 | 2.10e+04 | — | 100 | 5.01e+05 | — | 100 | 2.09e+04 | — | 100 | 1.25e+04 | — | 7.94 | 1.80e+06 | — | 65.77 | 2.75e+06 | — | 100 | 2.69e+04 | — |
| $f_{17}$ | — | 100 | 2.10e+04 | — | 100 | 5.01e+05 | — | 100 | 2.08e+04 | — | 100 | 1.23e+04 | — | 56.59 | 1.80e+06 | — | 98.93 | 1.73e+05 | — | 100 | 2.53e+04 | — |
| $f_{18}$ | — | 99.29 | 2.18e+04 | — | 100 | 5.01e+05 | — | 100 | 2.08e+04 | — | 100 | 1.26e+04 | — | 64.05 | 1.80e+06 | — | 61.48 | 1.56e+06 | — | 100 | 2.55e+04 | — |
| $f_{19}$ | — | 99.84 | 2.11e+04 | — | 100 | 5.01e+05 | — | 100 | 2.07e+04 | — | 100 | 1.25e+04 | — | 55.35 | 1.81e+06 | — | 99.92 | 1.37e+05 | — | 100 | 2.49e+04 | — |
| $f_{20}$ | — | 100 | 2.10e+04 | — | 100 | 5.01e+05 | — | 100 | 2.07e+04 | — | 100 | 1.36e+04 | — | 91.20 | 1.81e+06 | — | 65.75 | 2.70e+06 | — | 100 | 3.01e+04 | — |
| $f_{21}$ | — | 100 | 2.00e+03 | — | 100 | 5.01e+05 | — | 100 | 6.00e+03 | — | 100 | 4.01e+03 | — | 100 | 1.00e+05 | — | 0.00 | 5.54e+05 | — | 100 | 6.41e+03 | — |
| Avg. | 75.7 | 99.81 | 3.74e+05 | 43.27 | 100 | 5.01e+05 | 47.25 | 100 | 1.33e+04 | 60.0 | 100 | 1.19e+04 | 99.44 | 48.03 | 1.31e+06 | 100 | 54.38 | 6.21e+05 | 99.85 | 100 | 1.30e+05 | 4.71e–10 |

For nonseparable variables, SVG achieves 100% decomposition accuracy ($\rho_2$) on all the 16 functions containing nonseparable variables. DG and its variants also perform well on most of these functions. The reason is that if two variables are generally nonseparable, they must be additively nonseparable. As for VIL and FVIS, despite their competition on separable variables, they achieve very low decomposition accuracy on nonseparable variables. This is because, for a pair of nonseparable variables, the solutions sampled without care can hardly satisfy **Criterion 2**, which inevitably results in some omissions of interdependency.

To sum up, the six state-of-the-art algorithms merely show some superiority on separable variables or nonseparable ones, while SVG performs consistently well on both types of variables. This merit mainly profits from two factors: 1) The developed **Criterion 3** can correctly identify general separability beyond additive separability; 2) the algorithmic component TLPR can efficiently provide global optima of variables required by **Criterion 3**. For the detailed decomposition results of SVG, readers can refer to **Table S1** in the supplementary.

2) *Decomposition efficiency*: SVG shows distinctly different FE consumptions w.r.t. different function categories. On the functions with separable variables ($f_1$-$f_{15}$), the number of FEs it consumes varies in the range $[1 \times 10^5, 2 \times 10^5]$. The results are better than those of DG2 and VIL but worse than those of RDG and EVIID, whose number of FEs drops an order of magnitude on average. The mediocrity of SVG on these functions derives from the fact that it needs to frequently invoke TLPR to locate the global optima for separable variables. Fortunately, this cost is not in vain. The located global optima can be directly transferred to the optimization process of CC. The entry *dis* in **Table 2** demonstrates that these optima are very close to the real ones. This verifies the efficiency of TLPR on the one hand. On the other hand, this means that it is unnecessary to tackle separable variables in the optimization process of CC, which will save many FEs. We will further reveal this merit in the subsequent experiment.

As for the functions without separable variables ($f_{16}$-$f_{21}$), SVG achieves competitive decomposition efficiency. Its FE consumption is much fewer than those of DG2, FVIS, and VIL, and is similar to those of DG and RDG. SVG shows some inferiority when compared with EVIID which is an efficient decomposition algorithm developed most recently. Nevertheless, the numbers of FEs they consumed still share the same order of magnitude.

4.3 Investigation of optimization performance

To investigate the capability of SVG in enhancing the optimization performance of CC, we incorporated it into the DECC framework [50] and compared its final optimization results with those obtained by the above six competitors. DECC is a classic CC framework. It takes an improved DE algorithm called SaNSDE [51] as the sub-problem optimizer and optimizes all the sub-problems in a round-robin fashion.

During the experiment, the parameters of DECC were strictly set following the original paper. According to the guidelines in [44], the maximum number of allowed FEs was set to $3.0 \times 10^6$, covering the FEs required by both the decomposition and optimization processes. On each benchmark function, we independently ran each DECC algorithm 25 times and assessed its performance in terms of the median, mean, and standard deviation of the obtained optima. For statistical analysis, we first used the Kruskal-Wallis nonparametric one-way ANOVA test with a confidence interval of 0.95 to determine whether there is at least one method showing distinct optimization performance and then conducted a series of two-tailed Wilcoxon rank-sum tests at a significance level of 0.05 in a pairwise fashion to compare the optimization results. **Table 3** presents the final optimization results of the seven DECC algorithms, where the entries highlighted in bold represent the best optimization results. According to **Table 3**, we can make the following observations:

On the fully separable functions $f_1$-$f_5$, DECC-SVG shows excellent optimization performance. It obtains the optimal or near-optimal solution for each function and significantly outperforms its six competitors. It is worth mentioning that for $f_1$-$f_3$ and $f_5$, DECC-SVG directly inherits their optimal solutions located by SVG and does not consume any FE in the optimization process. As for $f_4$, although SVG misjudges a few variables to be nonseparable, the resulting sub-problem can be effortlessly solved by the optimizer SaNSDE for its low dimension (see **Table S1** in the supplementary).

DECC-SVG also shows prominent superiority on the partially separable functions $f_6$-$f_{15}$ involving separable variables. For most of these benchmark functions, DECC-SVG wins each of its six competitors by multiple orders of magnitude in terms of both median and mean. Especially, it generates near-optimal solutions for $f_8$, $f_{10}$, $f_{13}$, and $f_{15}$, while the solutions obtained by each of the other six DECC algorithms are far from the optima. As the seven DECC algorithms take the same CC framework and optimizer, the success of DECC-SVG mainly profits from the high decomposition accuracy and efficiency of SVG, which enable DECC-SVG to tackle nonseparable variables with sufficient FEs.

As for the partially separable functions $f_{16}$-$f_{20}$ that involve no separable variable and the non-separable one $f_{21}$, DECC-SVG also achieves competitive results in comparison with its six competitors. It shows almost the same optimization performance as DECC-DG, DECC-RDG, and DECC-EVIID. Such a result is readily comprehensible since the decomposition performance of SVG, DG, RDG, and EVIID on these functions differs slightly.

**Table 3.** Optimization results of the seven DECC algorithms on 1000-dimensional benchmark functions when $3 \times 10^6$ FEs are allowed.

| Func. | Stats | DECC-DG | DECC-DG2 | DECC-RDG | DECC-EVIID | DECC-VIL | DECC-FVIS | DECC-SVG |
|---|---|---|---|---|---|---|---|---|
| $f_1$ | Median | 3.48e+06 | 2.09e+06 | 9.24e+05 | 1.13e+06 | 1.51e+06 | 1.66e+06 | — |
|  | Mean | 1.22e+07 | 4.90e+06 | 1.78e+06 | 3.78e+06 | 3.33e+06 | 1.73e+06 | **0.00e+00** |
|  | Std | 2.30e+07 | 8.48e+06 | 2.77e+06 | 8.75e+06 | 5.74e+06 | 1.23e+06 | — |
| $f_2$ | Median | 7.54e+03 | 7.22e+03 | 6.73e+03 | 6.75e+03 | 6.84e+03 | 6.81e+03 | — |
|  | Mean | 7.57e+03 | 7.28e+03 | 6.75e+03 | 6.81e+03 | 6.84e+03 | 6.85e+03 | **0.00e+00** |
|  | Std | 2.23e+02 | 3.11e+02 | 3.22e+02 | 3.88e+02 | 1.97e+02 | 2.86e+02 | — |
| $f_3$ | Median | 1.66e+00 | 1.01e+00 | 9.35e-01 | 6.23e-01 | 8.67e-01 | 8.41e-01 | — |
|  | Mean | 2.14e+00 | 1.09e+00 | 1.08e+00 | 8.53e-01 | 9.29e-01 | 9.60e-01 | **3.23e-09** |
|  | Std | 1.30e+00 | 6.14e-01 | 8.49e-01 | 7.17e-01 | 5.59e-01 | 7.11e-01 | — |
| $f_4$ | Median | 1.11e+01 | 1.08e+01 | 1.06e+01 | 1.04e+01 | 2.04e+01 | 1.06e+01 | **1.64e-11** |
|  | Mean | 1.12e+01 | 1.09e+01 | 1.06e+01 | 1.04e+01 | 2.04e+01 | 1.07e+01 | **1.64e-11** |
|  | Std | 5.99e-01 | 8.36e-01 | 5.75e-01 | 6.24e-01 | 6.99e-02 | 7.94e-01 | **0.00e+00** |
| $f_5$ | Median | 1.86e+04 | 2.49e+04 | 1.32e+04 | 1.48e+04 | 1.80e+04 | 1.73e+04 | — |
|  | Mean | 2.31e+04 | 3.59e+04 | 1.74e+04 | 2.27e+04 | 2.28e+04 | 2.01e+04 | **8.49e-10** |
|  | Std | 1.64e+04 | 2.85e+04 | 1.44e+04 | 2.24e+04 | 1.89e+04 | 1.22e+04 | — |
| $f_6$ | Median | 2.06e+11 | 2.90e+10 | 2.47e+10 | 2.36e+10 | 1.89e+13 | 1.02e+13 | **8.16e+09** |
|  | Mean | 2.16e+11 | 3.66e+10 | 3.00e+10 | 3.11e+10 | 2.02e+13 | 1.17e+13 | **9.42e+09** |
|  | Std | 1.14e+11 | 2.47e+10 | 2.44e+10 | 2.46e+10 | 8.56e+12 | 4.95e+12 | **5.74e+09** |
| $f_7$ | Median | **1.20e+08** | **1.19e+08** | **1.10e+08** | **1.10e+08** | 4.92e+08 | **1.13e+08** | 1.15e+08 |
|  | Mean | **1.22e+08** | **1.16e+08** | **1.14e+08** | **1.16e+08** | 4.91e+08 | **1.20e+08** | 1.17e+08 |
|  | Std | **2.13e+07** | **2.10e+07** | **1.90e+07** | 3.01e+07 | 8.44e+07 | **2.63e+07** | 2.08e+07 |
| $f_8$ | Median | 5.47e+00 | 6.00e+00 | 4.16e+00 | 3.11e+00 | 4.38e+01 | 6.39e+01 | **4.86e-07** |
|  | Mean | 5.51e+00 | 6.24e+00 | 4.22e+00 | 3.01e+00 | 4.38e+01 | 6.52e+01 | **4.70e-07** |
|  | Std | 9.60e-01 | 1.26e+00 | 1.07e+00 | 9.72e-01 | 3.67e+00 | 6.72e+00 | **2.10e-07** |
| $f_9$ | Median | **1.73e+06** | **2.08e+06** | **2.14e+06** | **1.88e+06** | 2.07e+07 | **1.88e+06** | **1.95e+06** |
|  | Mean | **1.80e+06** | **2.08e+06** | **2.17e+06** | **1.74e+06** | 2.05e+07 | **1.75e+06** | **1.82e+06** |
|  | Std | **4.21e+05** | **4.35e+05** | **5.18e+05** | 8.15e+05 | **4.84e+05** | 8.19e+05 | 6.48e+05 |
| $f_{10}$ | Median | 2.01e+05 | 8.17e+04 | 1.28e+05 | 7.10e+04 | 2.17e+05 | 4.32e+09 | **3.09e-09** |
|  | Mean | 1.99e+05 | 8.67e+04 | 1.20e+05 | 7.70e+04 | 2.19e+05 | 7.70e+09 | **3.09e-09** |
|  | Std | 3.25e+04 | 2.47e+04 | 2.82e+04 | 3.13e+04 | 3.07e+04 | 6.31e+09 | **1.46e-21** |
| $f_{11}$ | Median | 1.13e+07 | 1.51e+07 | 8.20e+06 | 7.94e+06 | 7.72e+09 | 1.11e+10 | **3.07e+06** |
|  | Mean | 1.47e+07 | 1.64e+07 | 9.16e+06 | 1.22e+07 | 7.56e+09 | 1.08e+10 | **3.02e+06** |
|  | Std | 1.19e+07 | 8.02e+06 | 3.30e+06 | 1.01e+07 | 9.20e+08 | 9.97e+08 | **4.55e+05** |
| $f_{12}$ | Median | 5.71e+03 | 5.84e+03 | 5.61e+03 | 5.59e+03 | 1.37e+04 | 6.19e+03 | **1.37e+03** |
|  | Mean | 5.69e+03 | 5.82e+03 | 5.63e+03 | 5.56e+03 | 1.36e+04 | 6.18e+03 | **1.38e+03** |
|  | Std | 1.49e+02 | 1.80e+02 | 1.42e+02 | 1.59e+02 | 3.57e+02 | 1.71e+02 | **6.71e+01** |
| $f_{13}$ | Median | 5.33e+00 | 6.68e+00 | 4.85e+00 | 5.66e+00 | 8.31e+02 | 7.76e+02 | **1.58e-05** |
|  | Mean | 5.58e+00 | 6.78e+00 | 4.98e+00 | 5.63e+00 | 8.32e+02 | 7.80e+02 | **1.63e-05** |
|  | Std | 1.27e+00 | 1.66e+00 | 1.59e+00 | 1.64e+00 | 2.12e+01 | 3.56e+01 | **5.08e-06** |
| $f_{14}$ | Median | 2.93e+01 | 3.03e+01 | 3.02e+01 | 2.87e+01 | 2.29e+02 | 2.95e+01 | **2.01e+01** |
|  | Mean | 2.92e+01 | 3.01e+01 | 2.97e+01 | 2.87e+01 | 2.29e+02 | 2.93e+01 | **1.94e+01** |
|  | Std | 2.60e+00 | 2.49e+00 | 2.74e+00 | 2.30e+00 | 4.50e-01 | 2.72e+00 | **2.22e+00** |
| $f_{15}$ | Median | 3.34e+04 | 4.47e+04 | 3.22e+04 | 3.50e+04 | 3.96e+05 | 1.26e+06 | **3.02e-08** |
|  | Mean | 3.49e+04 | 4.38e+04 | 3.32e+04 | 3.49e+04 | 3.94e+05 | 1.28e+06 | **3.53e-07** |
|  | Std | 8.90e+03 | 1.07e+04 | 9.10e+03 | 1.07e+04 | 3.73e+04 | 8.11e+04 | **1.54e-06** |
| $f_{16}$ | Median | **1.44e+07** | 1.78e+07 | **1.49e+07** | **1.40e+07** | 1.99e+09 | 2.37e+11 | **1.49e+07** |
|  | Mean | **1.46e+07** | 1.84e+07 | **1.49e+07** | **1.46e+07** | 1.97e+09 | 2.31e+11 | **1.45e+07** |
|  | Std | **1.17e+06** | 1.57e+06 | **1.62e+06** | **1.82e+06** | 1.77e+08 | 3.23e+10 | **1.47e+06** |
| $f_{17}$ | Median | **3.37e+03** | 3.57e+03 | **3.41e+03** | **3.34e+03** | 1.61e+04 | 3.77e+03 | **3.39e+03** |
|  | Mean | **3.36e+03** | 3.57e+03 | **3.39e+03** | **3.33e+03** | 1.61e+04 | 3.81e+03 | **3.40e+03** |
|  | Std | **1.14e+02** | **1.12e+02** | **1.12e+02** | 1.59e+02 | 5.16e+02 | 1.51e+02 | **1.24e+02** |
| $f_{18}$ | Median | 2.12e-01 | 3.47e-01 | **9.76e-02** | **9.61e-02** | 2.21e+03 | 2.58e+03 | **9.50e-02** |
|  | Mean | 2.14e-01 | 3.47e-01 | **9.36e-02** | **9.70e-02** | 2.22e+03 | 2.59e+03 | **9.75e-02** |
|  | Std | 3.48e-02 | 6.95e-02 | **1.82e-02** | **1.25e-02** | 3.82e+01 | 3.01e+01 | **2.47e-02** |
| $f_{19}$ | Median | **3.97e+01** | **3.90e+01** | **3.77e+01** | **3.87e+01** | 4.16e+02 | **3.98e+01** | 4.01e+01 |
|  | Mean | **3.91e+01** | **3.85e+01** | **3.80e+01** | **3.91e+01** | 4.16e+02 | **3.97e+01** | 4.00e+01 |
|  | Std | **3.43e+00** | **2.09e+00** | **2.53e+00** | **2.91e+00** | 5.66e-01 | **2.46e+00** | 2.19e+00 |
| $f_{20}$ | Median | **1.51e-01** | 1.43e+00 | **1.41e-01** | **1.36e-01** | 2.20e+05 | 8.26e+07 | **1.43e-01** |
|  | Mean | **1.85e-01** | 1.59e+00 | **1.97e+00** | **2.44e-01** | 2.17e+05 | 8.55e+07 | 1.61e+00 |
|  | Std | **1.28e-01** | 7.08e-01 | **8.93e+00** | **2.43e-01** | 2.75e+04 | 2.76e+07 | 4.83e+00 |
| $f_{21}$ | Median | **8.82e+05** | 9.83e+05 | **8.71e+05** | **8.60e+05** | **8.77e+05** | 9.69e+05 | **8.56e+05** |
|  | Mean | **8.72e+05** | 1.01e+06 | **8.75e+05** | **8.73e+05** | 9.02e+05 | 9.79e+05 | **8.53e+05** |
|  | Std | **9.17e+04** | 9.68e+04 | **6.83e+04** | **6.47e+04** | 9.06e+04 | 7.96e+04 | **6.42e+04** |
| **No. of the Best** |  | 7 | 3 | 8 | 8 | 1 | 2 | 20 |

The bottom row of **Table 3** summarizes the number of functions on which each DECC algorithm achieves the best solution. It can be seen that DECC-SVG performs best on 20 out of all 21 benchmark functions and gets an apparent edge over the other six algorithms. DECC-RDG and DECC-EVIID both exhibit best optimization performance on 8 out of 21 functions and can be ranked second. They mainly win their scores on $f_{16}$-$f_{21}$, where RDG and EVIID achieve excellent decomposition accuracy and efficiency. Compared with the above two algorithms, DECC-DG loses its dominance on a function, i.e. $f_{18}$, where DG gets an improper decomposition. Due to the high FE requirement of DG2, DECC-DG2 shows unsatisfying performance on most functions and only get competitive solutions for $f_7$, $f_9$, and $f_{19}$.

As for DECC-VIL and DECC-FVIS, they perform poorly on most benchmark functions. Their failure on partially separable and nonseparable ones stems from the low capability of VIL and FVIS in identifying nonseparable variables. For fully separable functions, DECC-VIL and DECC-FVIS still perform mediocrely and generate almost the same optimization results as all the other algorithms except DECC-SVG. Such a result is somewhat counterintuitive since VIL and FVIS correctly identify most of the separable variables while DG and its variants misjudge some of them to be nonseparable. The reason consists in that now it is still an open issue to reasonably group separable variables to achieve efficient optimization, and the DECC framework simply treats all the separable variables as a whole. SVG handily avoids this puzzle by optimizing each separable variable with the TLPR scheme.

The above results indicate the superiority of SVG in enhancing the optimization performance of DECC when $3\times10^6$ FEs are allowed. Nevertheless, the available FEs for an LSGO problem are usually limited in practice. To further reveal the capability of SVG, we reduced the maximum number of allowed FEs to $2\times10^6$ and $1\times10^6$, and repeated the optimization experiment. Based on the detailed results (see **Table S2** and **Table S3** in the supplementary), we ranked the seven DECC algorithms according to the two-tailed Wilcoxon rank-sum test on each benchmark function. **Fig. 4** shows the final rankings with two radar charts, from which it can be seen that DECC-SVG still significantly outperforms its six competitors. It is ranked first on 20 and 19 out of all 21 functions when $2\times10^6$ and $1\times10^6$ FEs are allowed, respectively.

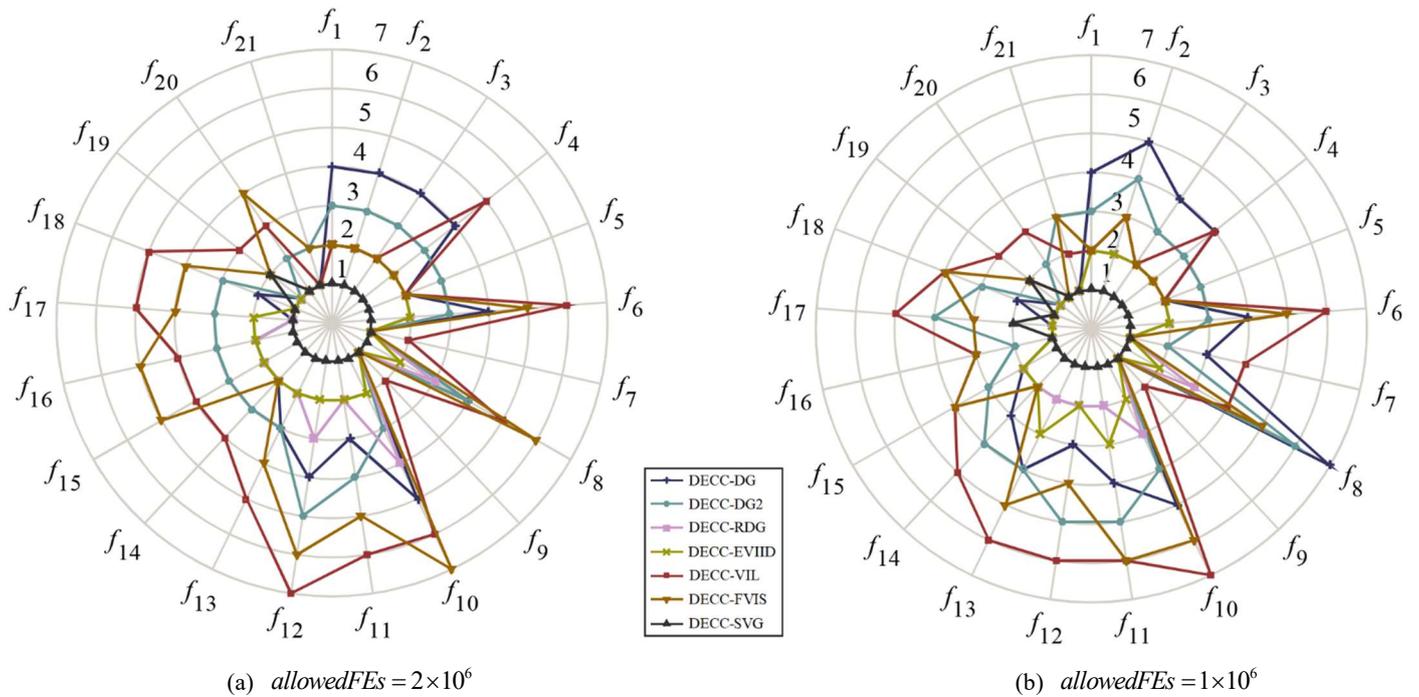

(a) $allowedFEs = 2\times10^6$        (b) $allowedFEs = 1\times10^6$

**Fig. 4.** Rankings of the seven DECC algorithms according to the two-tailed Wilcoxon rank-sum test on each 1000-dimensional benchmark function when two different maximum numbers of FEs are allowed.

## 4.4 Scalability studies

This experiment aims to study the scalability of SVG. To this end, we scaled the originally designed benchmark functions up to 2000 dimensions by resetting $n$ and $m$ to 2000 and 100, respectively, and re-conducted the above two sets of experiments on the new benchmark functions. Note that the maximum number of allowed FEs in this experiment was also doubled to $6.0 \times 10^6$.

**Table 4** provides the decomposition results of DG, DG2, RDG, EVIID, VIL, FVIS, and SVG. It can be observed that each of them achieves similar decomposition accuracy with that on 1000-dimensional functions. Especially, SVG correctly decomposes all the functions except for misjudging very few separable variables in $f_4$, $f_9$, and $f_{14}$, and shows a visible edge over its six competitors. For the detailed grouping results, readers can refer to **Table S4** in the supplementary. As for the decomposition efficiency, the seven algorithms increase their FE consumptions by different degrees. By comparing the bottom row of **Tables 2** and **4**, it can be known that the numbers of FEs consumed by DG, DG2, and FVIS increase by an order of magnitude, and those consumed by RDG, EVIID, VIL, and SVG keep the same order of magnitude. A closer observation indicates that SVG just doubles its number of FEs and demonstrates good scalability in problem decomposition.

**Table 5** reports the final optimization results of the seven DECC algorithms. It can be seen that DECC-SVG obtains the best solutions for 18 out of all 21 functions and significantly outperforms DECC-DG, DECC-DG2, DECC-RDG, DECC-EVIID, DECC-VIL, and DECC-FVIS, which obtain the best solutions for 5, 2, 6, 8, 0, and 2 functions, respectively. For functions on which DECC-SVG shows superiority, it either achieves near-optimal solutions or outperforms most of its competitors by several orders of magnitude in terms of both median and mean. For functions on which DECC-SVG shows some inferiority, it still shares the same order of magnitude with the corresponding winners. Therefore, SVG also dramatically enhances the optimization performance of DECC on 2000-dimensional problems.

**Table 4.** Decomposition results of the seven decomposition algorithms on the 2000-dimensional benchmark functions.

| Func. | DG | | | DG2 | | | RDG | | | EVIID | | | VIL | | | FVIS | | | SVG | | | |
|---|---|---|---|---|---|---|---|---|---|---|---|---|---|---|---|---|---|---|---|---|---|---|
| | $\rho_1$ | $\rho_2$ | FEsUsed | $\rho_1$ | $\rho_2$ | FEsUsed | $\rho_1$ | $\rho_2$ | FEsUsed | $\rho_1$ | $\rho_2$ | FEsUsed | $\rho_1$ | $\rho_2$ | FEsUsed | $\rho_1$ | $\rho_2$ | FEsUsed | $\rho_1$ | $\rho_2$ | FEsUsed | dis |
| $f_1$ | 100 | — | 4.00e+06 | 100 | — | 2.00e+06 | 100 | — | 6.01e+03 | 100 | — | 6.01e+03 | 100 | — | 1.32e+05 | 100 | — | 8.00e+04 | 100 | — | 4.22e+05 | 0.00e+00 |
| $f_2$ | 100 | — | 4.00e+06 | 100 | — | 2.00e+06 | 100 | — | 6.01e+03 | 100 | — | 6.01e+03 | 100 | — | 1.32e+05 | 100 | — | 8.00e+04 | 100 | — | 4.19e+05 | 2.33e-14 |
| $f_3$ | 100 | — | 4.00e+06 | 0.00 | — | 2.00e+06 | 0.00 | — | 1.25e+04 | 0.00 | — | 8.07e+03 | 100 | — | 1.32e+05 | 100 | — | 8.00e+04 | 100 | — | 3.93e+05 | 5.80e-05 |
| $f_4$ | 100 | — | 4.00e+06 | 0.00 | — | 2.00e+06 | 0.00 | — | 1.20e+04 | 0.00 | — | 8.65e+03 | 99.72 | — | 3.60e+06 | 100 | — | 8.82e+04 | 99.84 | — | 3.98e+05 | 3.36e-10 |
| $f_5$ | 0.23 | — | 4.01e+03 | 0.00 | — | 2.00e+06 | 0.00 | — | 1.20e+04 | 0.00 | — | 8.05e+03 | 100 | — | 1.32e+05 | 100 | — | 8.00e+04 | 100 | — | 4.02e+05 | 1.47e-12 |
| $f_6$ | 37.47 | 100 | 3.24e+04 | 100 | 100 | 2.00e+06 | 100 | 100 | 8.33e+03 | 100 | 100 | 6.13e+03 | 95.43 | 0.00 | 3.61e+06 | 100 | 0.00 | 7.86e+05 | 100 | 100 | 4.05e+05 | 0.00e+00 |
| $f_7$ | 100 | 100 | 3.62e+06 | 100 | 100 | 2.00e+06 | 100 | 100 | 8.36e+03 | 100 | 100 | 6.13e+03 | 97.18 | 0.00 | 3.61e+06 | 100 | 0.00 | 9.70e+04 | 100 | 100 | 4.01e+05 | 2.37e-14 |
| $f_8$ | 100 | 100 | 3.62e+06 | 9.73 | 100 | 2.00e+06 | 23.65 | 100 | 1.19e+05 | 100 | 100 | 3.49e+04 | 99.32 | 0.00 | 3.60e+06 | 100 | 0.00 | 3.71e+05 | 100 | 100 | 3.83e+05 | 7.85e-10 |
| $f_9$ | 100 | 100 | 3.62e+06 | 35.70 | 100 | 2.00e+06 | 100 | 100 | 5.92e+04 | 100 | 100 | 9.87e+03 | 97.63 | 0.00 | 3.60e+06 | 100 | 100 | 1.01e+05 | 99.95 | 100 | 3.84e+05 | 1.00e+00 |
| $f_{10}$ | 10.60 | 100 | 5.57e+03 | 65.44 | 100 | 2.00e+06 | 100 | 100 | 6.64e+05 | 100 | 100 | 3.58e+04 | 100 | 0.00 | 3.61e+06 | 100 | 0.00 | 1.15e+05 | 100 | 100 | 3.86e+05 | 3.40e-12 |
| $f_{11}$ | 100 | 100 | 1.05e+06 | 100 | 100 | 2.00e+06 | 100 | 100 | 2.81e+04 | 100 | 100 | 1.51e+04 | 99.14 | 49.32 | 3.60e+06 | 100 | 53.44 | 7.14e+06 | 100 | 100 | 2.41e+05 | 0.00e+00 |
| $f_{12}$ | 100 | 100 | 1.05e+06 | 100 | 100 | 2.00e+06 | 100 | 100 | 2.81e+04 | 100 | 100 | 1.69e+04 | 99.14 | 49.65 | 3.60e+06 | 100 | 97.10 | 2.55e+05 | 100 | 100 | 2.37e+05 | 1.67e-14 |
| $f_{13}$ | 100 | 87.75 | 1.13e+06 | 0.00 | 100 | 2.00e+06 | 0.00 | 100 | 2.66e+04 | 0.00 | 100 | 1.69e+04 | 99.70 | 51.64 | 3.60e+06 | 100 | 50.37 | 2.28e+06 | 100 | 100 | 2.25e+05 | 1.94e-05 |
| $f_{14}$ | 100 | 99.88 | 1.05e+06 | 0.00 | 100 | 2.00e+06 | 0.00 | 100 | 2.66e+04 | 0.00 | 100 | 1.42e+04 | 98.57 | 47.24 | 3.60e+06 | 100 | 100 | 2.21e+05 | 99.56 | 100 | 2.24e+05 | 1.52e-10 |
| $f_{15}$ | 0.23 | 100 | 1.100e+04 | 0.00 | 100 | 2.00e+06 | 0.00 | 100 | 2.65e+04 | 0.00 | 100 | 1.51e+04 | 100 | 87.45 | 3.60e+06 | 100 | 53.37 | 7.94e+06 | 100 | 100 | 2.31e+05 | 5.68e-13 |
| $f_{16}$ | — | 100 | 4.20e+04 | — | 100 | 2.00e+06 | — | 100 | 4.18e+04 | — | 100 | 2.52e+04 | — | 27.11 | 3.60e+06 | — | 60.16 | 1.36e+07 | — | 100 | 4.80e+04 | — |
| $f_{17}$ | — | 100 | 4.20e+04 | — | 100 | 2.00e+06 | — | 100 | 4.19e+04 | — | 100 | 2.48e+04 | — | 51.63 | 3.61e+06 | — | 97.79 | 3.97e+05 | — | 100 | 4.63e+04 | — |
| $f_{18}$ | — | 91.81 | 6.46e+04 | — | 100 | 2.00e+06 | — | 100 | 4.15e+04 | — | 100 | 2.53e+04 | — | 57.86 | 3.61e+06 | — | 56.91 | 4.49e+06 | — | 100 | 4.58e+04 | — |
| $f_{19}$ | — | 99.38 | 4.24e+04 | — | 100 | 2.00e+06 | — | 100 | 4.18e+04 | — | 100 | 2.48e+04 | — | 52.62 | 3.60e+06 | — | 100 | 2.77e+05 | — | 100 | 4.58e+04 | — |
| $f_{20}$ | — | 100 | 4.20e+04 | — | 100 | 2.00e+06 | — | 100 | 4.18e+04 | — | 100 | 2.79e+04 | — | 90.11 | 3.61e+06 | — | 60.17 | 1.70e+07 | — | 100 | 5.12e+04 | — |
| $f_{21}$ | — | 100 | 4.00e+03 | — | 100 | 2.00e+06 | — | 100 | 1.20e+04 | — | 100 | 8.00e+03 | — | 100 | 1.95e+05 | — | 0.00 | 1.79e+07 | — | 100 | 1.24e+04 | — |
| Avg. | 76.57 | 98.68 | 1.50e+06 | 47.39 | 100 | 2.00e+06 | 54.91 | 100 | 6.02e+04 | 60 | 100 | 1.64e+04 | 99.06 | 41.54 | 2.78e+06 | 100 | 51.83 | 3.49e+06 | 99.96 | 100 | 2.57e+05 | 6.66e-02 |

**Table 5.** Optimization results of the seven resulting DECC algorithms on the 2000-dimensional benchmark functions.

| Func. | Stats. | DECC-DG | DECC-DG2 | DECC-RDG | DECC-EVIID | DECC-VIL | DECC-FVIS | DECC-SVG |
|---|---|---|---|---|---|---|---|---|
| $f_1$ | Median | 5.07e+08 | 1.45e+08 | 4.24e+07 | 5.06e+07 | 4.74e+07 | 5.23e+07 | — |
|  | Mean | 5.22e+08 | 1.90e+08 | 5.89e+07 | 6.32e+07 | 6.18e+07 | 8.39e+07 | **0.00e+00** |
|  | Std | 1.63e+08 | 1.41e+08 | 5.18e+07 | 4.14e+07 | 3.80e+07 | 1.08e+08 | — |
| $f_2$ | Median | 2.18e+04 | 1.91e+04 | 1.76e+04 | 1.79e+04 | 1.77e+04 | 1.77e+04 | — |
|  | Mean | 2.16e+04 | 1.90e+04 | 1.75e+04 | 1.78e+04 | 1.77e+04 | 1.76e+04 | **0.00e+00** |
|  | Std | 5.93e+02 | 1.06e+03 | 9.34e+02 | 3.55e+02 | 5.39e+02 | 9.75e+02 | — |
| $f_3$ | Median | 9.78e-01 | 5.38e-01 | 3.34e-01 | 3.07e-01 | 3.28e-01 | 3.34e-01 | — |
|  | Mean | 9.61e-01 | 5.26e-01 | 3.35e-01 | 3.06e-01 | 3.31e-01 | 3.32e-01 | **2.44e-08** |
|  | Std | 8.28e-02 | 6.81e-02 | 8.30e-02 | 7.63e-02 | 7.20e-02 | 7.37e-02 | — |
| $f_4$ | Median | 1.33e+01 | 1.26e+01 | 1.22e+01 | 1.23e+01 | 2.07e+01 | 1.22e+01 | **3.01e-11** |
|  | Mean | 1.33e+01 | 1.27e+01 | 1.22e+01 | 1.23e+01 | 2.07e+01 | 1.22e+01 | **3.01e-11** |
|  | Std | 2.55e-01 | 3.56e-01 | 3.12e-01 | 2.73e-01 | 2.58e-02 | 3.11e-01 | **1.63e-15** |
| $f_5$ | Median | 7.30e+05 | 5.43e+05 | 3.24e+05 | 3.33e+05 | 3.57e+05 | 3.25e+05 | — |
|  | Mean | 7.30e+05 | 5.50e+05 | 3.08e+05 | 3.30e+05 | 3.57e+05 | 3.24e+05 | **2.93e-09** |
|  | Std | 7.50e+04 | 8.45e+04 | 7.13e+04 | 4.99e+04 | 6.05e+04 | 7.56e+04 | — |
| $f_6$ | Median | 1.48e+12 | 5.12e+11 | 3.25e+11 | 3.48e+11 | 5.22e+14 | 2.43e+13 | **2.01e+11** |
|  | Mean | 1.53e+12 | 5.77e+11 | 3.59e+11 | 3.79e+11 | 5.36e+14 | 2.48e+13 | **2.22e+11** |
|  | Std | 3.96e+11 | 2.97e+11 | 1.48e+11 | 1.22e+11 | 1.27e+14 | 6.80e+12 | **1.35e+11** |
| $f_7$ | Median | 3.21e+08 | 3.05e+08 | 2.96e+08 | **2.82e+08** | 1.43e+09 | 3.20e+08 | 3.15e+08 |
|  | Mean | 3.16e+08 | 3.15e+08 | 2.96e+08 | **2.81e+08** | 1.44e+09 | 3.20e+08 | 3.18e+08 |
|  | Std | 3.12e+07 | 3.71e+07 | 3.75e+07 | **3.92e+07** | 1.30e+08 | 3.94e+07 | 5.94e+07 |
| $f_8$ | Median | 2.11e+01 | 5.38e+00 | 3.39e+00 | 7.04e-01 | 2.13e+06 | 3.85e+01 | **4.33e-07** |
|  | Mean | 2.55e+01 | 8.53e+00 | 4.25e+00 | 6.99e-01 | 2.10e+06 | 5.09e+01 | **4.52e-07** |
|  | Std | 2.15e+01 | 8.97e+00 | 3.34e+00 | 8.14e-02 | 2.79e+05 | 3.85e+01 | **9.21e-08** |
| $f_9$ | Median | **3.93e+06** | 4.31e+06 | **3.80e+06** | **3.92e+06** | 2.08e+07 | **4.10e+06** | **3.89e+06** |
|  | Mean | **3.97e+06** | 4.16e+06 | **3.77e+06** | **3.83e+06** | 2.08e+07 | **4.11e+06** | **3.95e+06** |
|  | Std | **5.04e+05** | 9.25e+05 | **7.31e+05** | **7.16e+05** | 7.48e+04 | **6.98e+05** | **9.48e+05** |
| $f_{10}$ | Median | 1.05e+06 | 2.65e+05 | 6.42e+05 | 5.88e+05 | 1.48e+09 | 3.97e+06 | **6.46e-09** |
|  | Mean | 7.29e+06 | 2.75e+05 | 6.55e+05 | 6.05e+05 | 2.26e+09 | 1.42e+07 | **6.46e-09** |
|  | Std | 3.12e+07 | 5.84e+04 | 7.72e+04 | 8.53e+04 | 2.19e+09 | 5.13e+07 | **1.17e-19** |
| $f_{11}$ | Median | 1.71e+08 | 2.72e+08 | 1.29e+08 | 1.18e+08 | 3.06e+10 | 9.63e+11 | **1.82e+07** |
|  | Mean | 2.00e+08 | 3.29e+08 | 1.43e+08 | 1.35e+08 | 3.08e+10 | 9.70e+11 | **1.82e+07** |
|  | Std | 9.15e+07 | 2.07e+08 | 6.08e+07 | 4.76e+07 | 2.68e+09 | 2.15e+10 | **1.77e+06** |
| $f_{12}$ | Median | 1.52e+04 | 1.58e+04 | 1.44e+04 | 1.45e+04 | 2.95e+04 | 1.64e+04 | **4.21e+03** |
|  | Mean | 1.51e+04 | 1.58e+04 | 1.44e+04 | 1.45e+04 | 2.96e+04 | 1.64e+04 | **4.19e+03** |
|  | Std | 2.48e+02 | 2.41e+02 | 3.04e+02 | 3.63e+02 | 4.82e+02 | 2.34e+02 | **1.90e+02** |
| $f_{13}$ | Median | 3.31e+00 | 1.53e+00 | 1.12e+00 | 1.04e+00 | 9.93e+01 | 8.32e+01 | **2.61e-02** |
|  | Mean | 3.27e+00 | 1.57e+00 | 1.13e+00 | 1.03e+00 | 9.92e+01 | 8.38e+01 | **2.80e-02** |
|  | Std | 1.94e-01 | 1.29e-01 | 1.43e-01 | 9.60e-02 | 1.83e+00 | 2.10e+00 | **1.12e-02** |
| $f_{14}$ | Median | 5.40e+01 | 5.43e+01 | 5.35e+01 | 5.29e+01 | 2.29e+02 | 5.34e+01 | **4.08e+01** |
|  | Mean | 5.38e+01 | 5.45e+01 | 5.34e+01 | 5.33e+01 | 2.29e+02 | 5.36e+01 | **4.05e+01** |
|  | Std | 3.58e+00 | 2.57e+00 | 2.73e+00 | 2.30e+00 | 2.64e-01 | 2.83e+00 | **2.06e+00** |
| $f_{15}$ | Median | 2.98e+05 | 3.90e+05 | 2.86e+05 | 2.86e+05 | 1.63e+06 | 2.62e+08 | **1.92e+01** |
|  | Mean | 3.03e+05 | 3.86e+05 | 2.91e+05 | 2.89e+05 | 1.64e+06 | 3.28e+08 | **8.59e+01** |
|  | Std | 2.88e+04 | 3.41e+04 | 3.85e+04 | 3.14e+04 | 8.12e+04 | 1.05e+08 | **1.27e+02** |
| $f_{16}$ | Median | **7.03e+07** | 1.11e+08 | **7.14e+07** | **7.17e+07** | 2.83e+10 | 9.19e+11 | **7.06e+07** |
|  | Mean | **7.05e+07** | 1.12e+08 | **7.11e+07** | **7.11e+07** | 2.81e+10 | 9.20e+11 | **7.12e+07** |
|  | Std | **4.72e+06** | 5.19e+06 | **4.33e+06** | **3.62e+06** | 2.30e+09 | 5.87e+10 | **3.82e+06** |
| $f_{17}$ | Median | **9.28e+03** | 1.06e+04 | **9.15e+03** | **9.25e+03** | 3.64e+04 | 1.17e+04 | 9.70e+03 |
|  | Mean | **9.25e+03** | 1.06e+04 | **9.20e+03** | **9.19e+03** | 3.63e+04 | 1.17e+04 | 9.76e+03 |
|  | Std | **2.78e+02** | 2.63e+02 | **2.37e+02** | **2.51e+02** | 9.44e+02 | 3.70e+02 | 3.16e+02 |
| $f_{18}$ | Median | 3.42e+00 | 1.54e+00 | **6.85e-01** | 7.04e-01 | 2.19e+02 | 2.53e+02 | **6.76e-01** |
|  | Mean | 3.38e+00 | 1.53e+00 | **7.03e-01** | 7.03e-01 | 2.19e+02 | 2.53e+02 | **6.95e-01** |
|  | Std | 2.88e-01 | 1.68e-01 | **9.49e-02** | 8.23e-02 | 2.08e+00 | 1.87e+00 | **1.10e-01** |
| $f_{19}$ | Median | **8.12e+01** | **8.09e+01** | **8.15e+01** | **8.02e+01** | 4.17e+02 | **8.15e+01** | **8.06e+01** |
|  | Mean | **8.09e+01** | **8.08e+01** | **8.15e+01** | **8.05e+01** | 4.17e+02 | **8.14e+01** | **8.07e+01** |
|  | Std | **3.14e+00** | **3.02e+00** | **3.49e+00** | 4.97e+00 | 3.47e-01 | **2.84e+00** | **3.69e+00** |
| $f_{20}$ | Median | **2.26e+03** | 1.35e+04 | **2.25e+03** | 2.66e+03 | 8.51e+05 | 6.04e+08 | **2.37e+03** |
|  | Mean | **2.43e+03** | 1.38e+04 | **2.38e+03** | 2.89e+03 | 8.46e+05 | 5.53e+08 | **2.97e+03** |
|  | Std | **1.38e+03** | 2.30e+03 | **7.67e+02** | 1.58e+03 | 5.99e+04 | 1.46e+08 | **2.34e+03** |
| $f_{21}$ | Median | 3.50e+06 | 4.28e+06 | 3.49e+06 | **3.39e+06** | 3.46e+06 | 7.50e+09 | 3.51e+06 |
|  | Mean | 3.51e+06 | 4.27e+06 | 3.52e+06 | **3.36e+06** | 3.48e+06 | 8.87e+09 | 3.52e+06 |
|  | Std | 2.95e+05 | 3.39e+05 | 2.45e+05 | **1.72e+05** | 2.63e+05 | 6.65e+09 | 2.37e+05 |
| **No. of the Best** |  | 5 | 2 | 6 | 8 | 0 | 2 | 18 |

# 5. Conclusion

In this paper, we present a novel decomposition algorithm named SVG for CC. This algorithm can efficiently decompose general LSGO problems with a high accuracy, which benefits from the new designed separability detection criterion. Checking whether the global optimum of a variable relies on other variables, this criterion is consistent with the definition of general separability and extends SVG beyond additive separability. To ensure its decomposition efficiency, SVG designs a two-layer polynomial regression scheme and a dynamic-binary-tree-based variable grouping procedure. The former can locate the global optimum of a variable required by the separability detection criterion with a small number of FEs, and the latter systematically reutilizes historical separability detection information and thus can reduce detection times. Extensive experimental results on a new designed benchmark suite for generally separable LSGO problems reveal that compared with six state-of-the-art decomposition algorithms, SVG owns impressive accuracy, efficiency, scalability, and capability in enhancing the optimization performance of CC.

It is worth mentioning that this study temporarily takes no account of indirect interdependency since now it is still an open issue to group indirectly nonseparable variables for possibly resulting high-dimensional sub-problems [42]. We will improve SVG to deal with this special case in our future work. Besides, we will further verify the performance of SVG by applying it to real-world LSGO problems.

# Credit authorship contribution statement

**An Chen**: Conceptualization, Methodology, Software, Validation, Investigation, Writing-original draft, Writing-review & editing. **Zhigang Ren**: Conceptualization, Methodology, Visualization, Investigation, Writing-review & editing. **Muyi Wang**: Software, Validation, Visualization, Investigation, Writing-review & editing. **Yongsheng Liang**: Software, Investigation, Writing-review & editing. **Hanqing Liu**: Writing-review & editing. **Wenhao Du**: Visualization, Writing-review & editing.

# Declaration of Competing Interest

The authors declare that they have no known competing financial interests or personal relationships that could have appeared to influence the work reported in this paper.

# Acknowledgements

This work was supported by the National Natural Science Foundation of China (grant number 61873199), the Natural Science Basic Research Plan in Shaanxi Province of China (grant number 2020JM-059), and the Fundamental Research Funds for the Central Universities [grant numbers xzy022020057]